\documentclass{article}

     \PassOptionsToPackage{numbers, compress}{natbib}

 \usepackage[main, final]{neurips_2026}

\usepackage[utf8]{inputenc} 
\usepackage[T1]{fontenc}    
\usepackage{hyperref}       
\usepackage{url}            
\usepackage{booktabs}       
\usepackage{amsfonts}       
\usepackage{nicefrac}       
\usepackage{microtype}      

\usepackage[table]{xcolor}
\usepackage{amsmath}
\usepackage{tabularx}
\usepackage{array}
\usepackage{graphicx}
\usepackage{multirow}
\usepackage{comment}
\usepackage{subcaption}
\usepackage{adjustbox}
\usepackage{bm} 
\usepackage{wrapfig}
\usepackage{algorithm}
\usepackage{algorithmic}

\newcommand{\vk}{\bm{k}}
\newcommand{\vs}{\bm{s}}

\definecolor{color3}{HTML}{F2F3F5}
\definecolor{color-ours}{HTML}{E8F7FF}
\definecolor{color-activation}{HTML}{5DC055}
\definecolor{color-weight}{HTML}{6CA7F5}

\title{QuantDemoire: Quantization with Outlier Aware for Image Demoir\'eing}

\author{Zheng Chen$^{1}$\thanks{Equal contribution.},\enspace 
Kewei Zhang$^{1}$\footnotemark[1],\enspace
Xiaoyang Liu$^{1}$,\enspace 
Kai Liu$^{1}$,\enspace \\
\textbf{Weihang Zhang$^{2}$,}\enspace 
\textbf{Mengfan Wang$^{2}$,}\enspace 
\textbf{Yifan Fu$^{2}$,}\enspace 
\textbf{Yulun Zhang$^{1}$}\thanks{Corresponding author: Yulun Zhang, yulun100@gmail.com} \\
\textsuperscript{1}Shanghai Jiao Tong University,\enspace
\textsuperscript{2}Institute of Media Technology and Experience Design, Huawei
\vspace{-5.mm}
}

\begin{document}

\maketitle

\begin{abstract}
Demoir\'eing aims to remove moir\'e artifacts that often occur in images. While recent deep learning-based methods have achieved promising results, they typically require substantial computational resources, limiting their deployment on edge devices. Model quantization offers a compelling solution. However, directly applying existing quantization methods to demoir\'eing models introduces severe performance degradation.  The main reasons are distribution outliers and weakened representations in smooth regions. To address these issues, we propose QuantDemoire, a post-training quantization framework tailored to demoir\'eing. It contains two key components. \textbf{First}, we introduce an outlier-aware quantizer to reduce errors from outliers. It uses sampling-based range estimation to reduce activation outliers and keeps a few extreme weights in FP16 with negligible cost. \textbf{Second}, we design a frequency-aware calibration strategy. It emphasizes low- and mid-frequency components during fine-tuning, which mitigates banding artifacts caused by low-bit quantization. Experiments validate that our QuantDemoire achieves large reductions in parameters and computation while maintaining quality. Meanwhile, it outperforms existing quantization methods by over \textbf{4 dB} on W4A4. We implement a CUDA kernel for the mixed-precision branch and achieve a 2.3$\times$ end-to-end speedup on an RTX 4090 GPU. Code is released at:~\url{https://github.com/zhengchen1999/QuantDemoire}.

\end{abstract}

\setlength{\abovedisplayskip}{2pt}
\setlength{\belowdisplayskip}{2pt}

\vspace{-6.mm}
\section{Introduction}
\vspace{-3.mm}
Image demoir\'eing is a low-level vision task that removes colored stripes caused by frequency aliasing between display screens and imaging sensors. These artifacts reduce image quality and hinder later analysis and recognition. Due to the complex and diverse forms of moir\'e patterns, effective demoir\'eing remains a highly challenging task~\cite{sidorov2002suppression,liu2015moire,wang2023coarse}.

In recent years, with the rapid progress of deep learning, neural networks have achieved remarkable performance in demoir\'eing~\cite{cheng2019multi,zheng2020image,yu2022towards,xu2023image,liu2025freqformer}. They adapt to the diverse moir\'e patterns and achieve excellent performance. However, they usually involve millions of parameters and heavy computation. This limits their use on devices with extremely limited computing power, such as drones, portable cameras and IoT sensors~\cite{cheng2019multi,zheng2020image}. These devices are also the most important application scenarios of image demoir\'eing. Therefore, efficient and high-quality demoir\'eing is critical for practical usage.

\begin{figure}[t]
\centering
\hspace{-0.mm}\includegraphics[width=\linewidth]{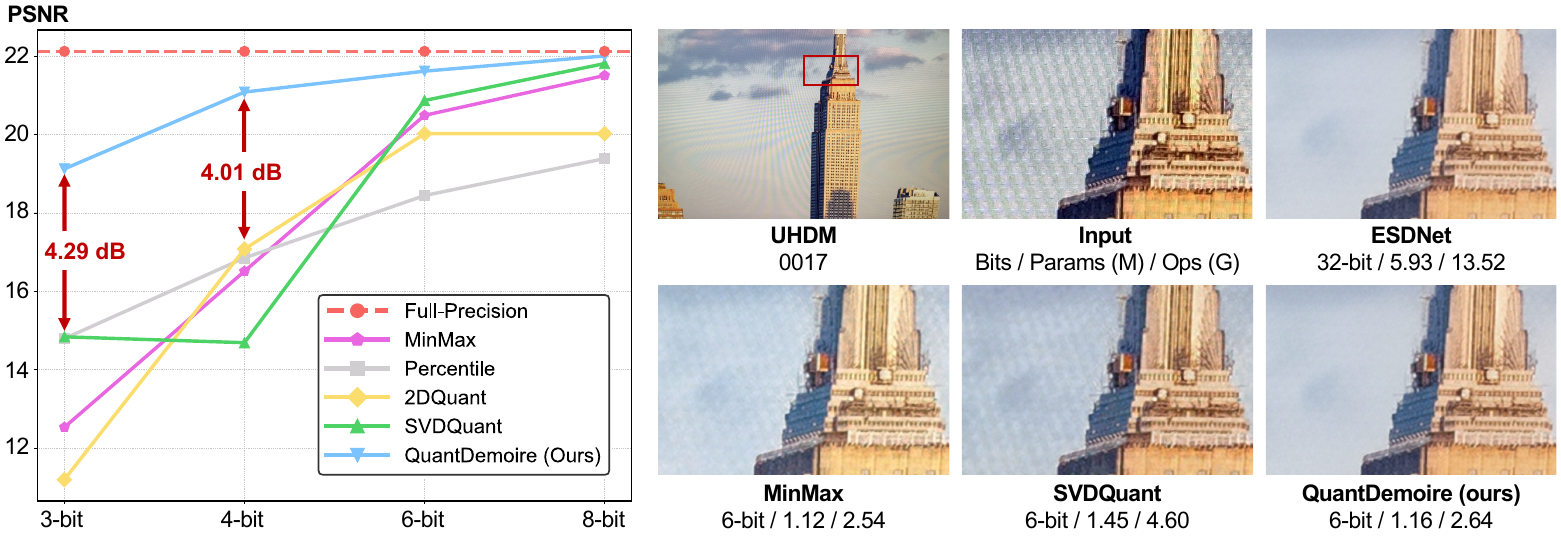} \\
\vspace{-1.mm}
\caption{Comparison of PSNR with other quantization methods on UHDM. The full-precision backbone is ESDNet~\cite{yu2022towards}. Our QuantDemoire demonstrates superior performance.}
\label{fig:performance_psnr}
\vspace{-5.mm}
\end{figure}

Quantization~\cite{nagel2021white,cheng2026permuquant,yan2026pt} is an effective solution to this challenge. It compresses weights and activations from 32-bit floats to low-bit (\textit{\textit{e.g.}}, 2$\sim$8 bit) integers. This significantly reduces storage overhead and replaces costly floating-point operations with efficient integer operations. Therefore, quantization speeds up inference and lowers power use, making it well-suited for resource-limited devices. 
Currently, quantization is widely applied to vision tasks~\cite{esser2019learned,li2023q,qin2023quantsr}.

However, demoir\'eing lacks dedicated quantization methods. {For extremely limited compute and memory, demoir\'eing models must be compressed to ultra-low bit widths (\textit{e.g.}, 4-bit). Image demoir\'eing is particularly sensitive to such quantization errors because it must suppress false high-frequency moir\'e patterns while preserving real textures. As a result, } directly applying existing quantization~\cite{jacob2017fakequant,li2024svdquant} results in severe performance loss. The degradation has two causes. \textbf{First}, weights and activations outliers expand the quantization range. This reduces the precision of typical values and weakens model's performance. \textbf{Second}, quantization weakens representations in smooth images regions. It introduces banding artifacts in mid- and low-frequency components. 
This not only harms demoir\'eing performance but also damages details in clean areas. 
Therefore, it is crucial to design task-specific quantization strategies tailored to the characteristics of the image demoir\'eing task.

To address these issues, we propose QuantDemoire, a post-training quantization (PTQ) framework for image demoir\'eing tasks. Our core idea reduces the accuracy loss via the quantizer and the calibration. \textbf{(1) Quantizer.} We design the outlier-aware quantizer to reduce the impact of outliers. It uses random sampling to estimate and remove activation outliers. It also keeps a tiny fraction of extreme weights in FP16, with negligible overhead. 
\textbf{(2) Calibration.} We introduce frequency-aware calibration. It emphasizes mid- and low-frequency components when optimizing quantization boundaries. This reduces the banding artifacts in smooth regions caused by quantization.

Extensive experiments show our QuantDemoire significantly improves efficiency while maintaining strong performance. It consistently outperforms existing quantization methods across datasets and bit-widths. As shown in Fig.~\ref{fig:performance_psnr}, at 4-bit, our method cuts parameters and computation by over \textbf{86.6\%} with under \textbf{4.7\%} PSNR drop vs. FP16. Besides, QuantDemoire surpasses SOTA quantization methods by over \textbf{4 dB} (PSNR, 4-bit) on UHDM~\cite{yu2022towards}.
Our main contributions are as follows:
\vspace{-1.mm}
\begin{itemize}
\item We propose QuantDemoire, the first quantization framework tailored to image demoir\'eing. It reduces model overhead while preserving restoration ability.
\vspace{-1.mm}
\item We design the outlier-aware quantizer that mitigates quantization errors by handling outliers in activations and weights with random sampling and FP16 branch. 
\vspace{-1.mm}
\item We propose the frequency-aware calibration strategy to optimize mid- and low-frequency features, alleviating quantization-induced banding artifacts.
\vspace{-1.mm}
\item We validate our method on multiple datasets and bit-widths. QuantDemoire consistently outperforms existing quantization methods quantitatively and visually.
\end{itemize}

\vspace{-4.mm}
\section{Related Work}
\vspace{-2.mm}

\subsection{Image demoir\'eing.} 
\vspace{-1.mm}
Photos of digital screens often contain colored moir\'e stripes. These patterns can greatly degrade image quality. Early moir\'e pattern removal methods used traditional mathematical methods, such as matrix decomposition~\cite{liu2015moire} and spectral models~\cite{sidorov2002suppression}. These techniques generally perform poorly when dealing with diverse types of moir\'e patterns. In recent years, with the rapid development of deep learning, methods~\cite{cheng2019multi,he2019mop,zheng2020image,he2020fhde2net,zhou2026bricker,yu2022towards,xu2023image,xiao2024p,liu2025freqformer}, based on deep neural networks,  have been proposed. MopNet~\cite{he2019mop} utilizes multi-scale feature aggregation and attribute-aware classifiers to handle complex frequency. 
FHDe$^2$Net~\cite{he2020fhde2net} employs a cascaded architecture to remove multi-scale moir\'e patterns. 
ESDNet~\cite{yu2022towards} adds a semantic-aligned, scale-aware module to further improve robustness to scale variations. However, these methods have millions of parameters. The high cost makes them unsuitable for edge devices with limited resources. 

\subsection{Model Quantization.}
\vspace{-2.mm}
Quantization is widely used in neural network compression. It effectively reduces memory and computation by converting 32-bit weights and activations to low-bit integers. Quantization algorithms are generally divided into quantization-aware training (QAT)~\cite{esser2019learned,efficientqat,qin2023quantsr} and post-training quantization (PTQ)~\cite{jacob2017fakequant,liu2021post,shang2023post,li2024svdquant,zhu2025passionsr,yan2026focus}. 
QAT can be accurate~\cite{nagel2021white}, but its high training cost limits practical use. PTQ is lightweight and efficient because it requires no retraining or labeled data. 
SmoothQuant~\cite{xiao2023smoothquant} shifts the quantization difficulty from activations to weights. 
SVDQuant~\cite{li2024svdquant} decomposes weights into a low-rank component.
Nevertheless, these quantization methods suffer from severe performance degradation on demoir\'eing models. The influence of outliers~\cite{dettmers2022gpt3} and information loss from quantization degrade quality, causing color banding and poor moir\'e removal. Appendix~\ref{sec:app_related_methods} further discusses the differences between QuantDemoire and smoothing or mixed-precision quantization methods.

\vspace{-4.mm}
\section{Method}
\vspace{-3.mm}
\label{headings}

\begin{figure*}[t]
\centering
\hspace{-0.mm}\includegraphics[width=1.\linewidth]{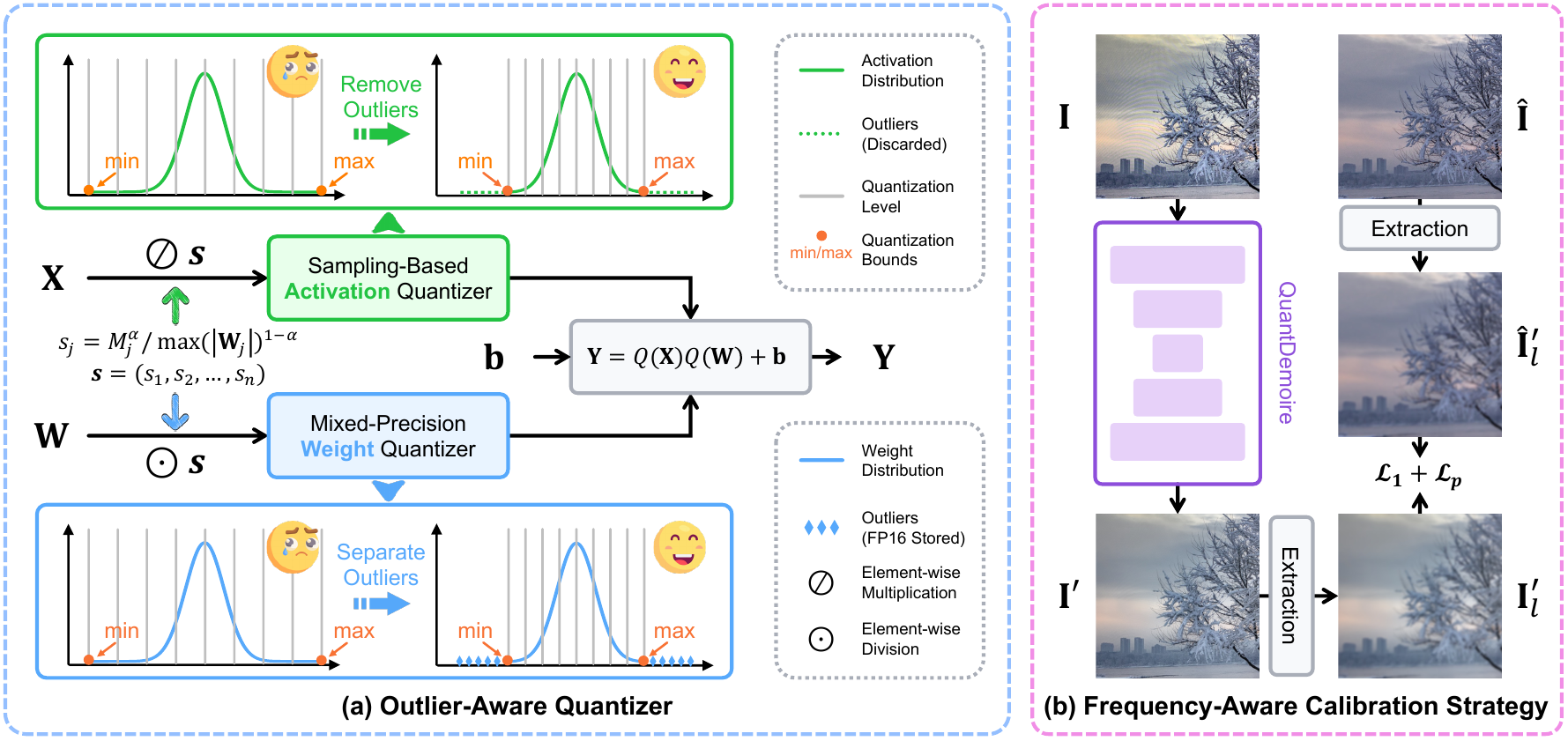} \\
\vspace{-1.mm}
\caption{The overview of our QuantDemoire. (a) Outlier-Aware Quantizer: activations discard outliers through sample-based estimation, while weights preserve outliers in FP16 precision. (b) Frequency-Aware Calibration Strategy: quantizer parameters are optimized on mid- and low-frequency components, which are extracted through the frequency extraction process.}
\label{fig:overview}
\vspace{-7.mm}
\end{figure*}

In this section, we present our quantization framework, QuantDemoire, for efficient demoir\'eing (see Fig.~\ref{fig:overview}). We introduce two key components. The first component is the outlier-aware quantizer, which reduces the quantization error introduced by outliers in activations and weights. The second is the frequency-aware calibration strategy, which selectively emphasizes low- and mid-frequency components to mitigate the banding artifacts during the calibration phase.

\vspace{-3.mm}
\subsection{Preliminaries: Quantization}
\vspace{-2.mm}
We first introduce fake quantization~\cite{jacob2017fakequant}, which is employed to simulate quantization errors. Given a weight or activation tensor value $v$ and a quantization bit-width $b$, the procedure is expressed as:
\begin{equation}
\begin{gathered}
\Delta = \frac{u-l}{2^b - 1}, \quad z = \text{clip}(\text{Round}(-\frac{l}{\Delta}),0,2^b -1), \\ \quad v_z = \text{clip}(\text{Round}(\frac{v}{\Delta}) + z,0,2^b -1),
\label{eq:quantize}
\end{gathered}
\end{equation}
where $l$ and $u$ denote the lower and upper quantization bounds, $\Delta$ is the scaling factor, $z$ is the zero-point offset, and $v_z$ is the integer-coded value of $v$. The clipping function is defined as $\text{clip}(v,l,u)=\text{max}(\text{min}(v,u),l)$, while $\text{Round}(\cdot)$ maps an input to its nearest integer. The corresponding dequantization process can be formulated as:
\begin{equation}
\begin{gathered}
Q(v) = \Delta \times (v_z -z),
\label{eq:dequantize}
\end{gathered}
\end{equation}
where $Q(\cdot)$ denotes the operation of fake quantization. Given the input activation $\mathbf{X}$, weight $\mathbf{W}$, and bias $\mathbf{b}$, the output feature $\mathbf{Y}$ can be obtained through:
\begin{equation}
\begin{gathered}
\mathbf{Y} = Q(\mathbf{X}) \cdot Q(\mathbf{W}) +\mathbf{b}.
\label{eq:quantconv}
\end{gathered}
\end{equation}
Because $\text{Round}(\cdot)$ is non-differentiable, following prior works~\cite{qin2023quantsr,li2024svdquant}, the straight-through estimator (STE) is adopted to approximate its gradient. The process is defined as:
\begin{equation}
\begin{gathered}
\frac{\partial Q(v)}{\partial v} \approx \begin{cases} 1 & \text{if } v \in [l, u], \\ 0 & \text{otherwise}. \end{cases}
\label{eq:STE}
\end{gathered}
\end{equation}
Moreover, SmoothQuant~\cite{xiao2023smoothquant} is widely adopted in quantization to address the challenge arising from the variation in maximum magnitudes across channels. It applies a per-channel scaling transformation to redistribute the quantization difficulty from activations to weights. Given an activation $\mathbf{X}$ and a weight tensor $\mathbf{W}$ for the $j$-th input channel, the smoothing factor $s_j$ is defined as:
\begin{equation}
\begin{gathered}
s_j = {\max(|\mathbf{X}_j|)^\alpha}/{\max(|\mathbf{W}_j|)^{1-\alpha}}, \\ \quad \mathbf{s} = (s_1,s_2,\dots,s_{n}),
\label{eq:smoothquant}
\end{gathered}
\end{equation}
where $\alpha$ is a balancing hyperparameter controlling the trade-off between activation and weights, $n$ denotes the number of input channels, and $\mathbf{s}$ is the smoothing scale vector.

Despite these developments, existing quantization methods still lead to substantial performance degradation when applied to demoir\'eing models. Recent methods are mainly designed for LLMs or DiTs and do not adequately address outliers in convolutional modules. Moreover, generic quantization approaches lack task-specific optimization tailored to demoir\'eing. Therefore, we propose the outlier-aware quantizer and the frequency-aware calibration strategy to enhance the performance.

\vspace{-3.mm}
\subsection{Outlier-aware quantizer}
\vspace{-2.mm}
\label{sec:quantizer}
\noindent \textbf{Challenges.} Despite the progress in neural network quantization, existing methods still face critical limitations, particularly due to the presence of outliers. They can lead to two key problems: \textbf{\textit{quantization range expansion}} and \textbf{\textit{inaccurate smoothing scale factor estimation}}.

\textbf{\textit{Challenge I. Quantization range expansion.}} Outliers enlarge the quantization range either within a single channel or across the entire tensor, thereby reducing information density of normal values and aggravating quantization error for them, especially under low-bit quantization.

\begin{figure*}[t]
\centering
\hspace{-0.mm}\includegraphics[width=1.\linewidth]{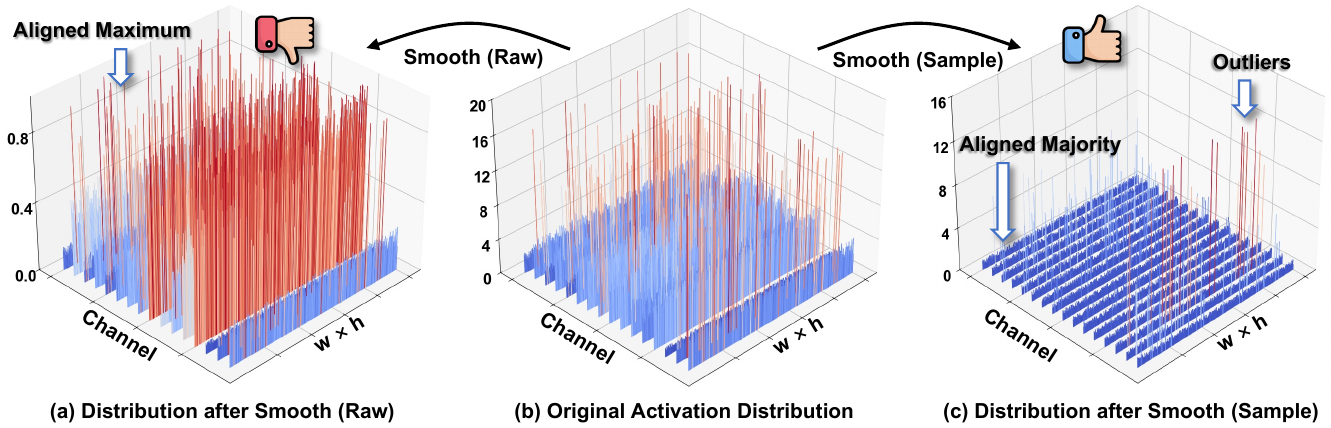} \\
\vspace{-1.mm}
\caption{Activation distributions (Original, Smooth (Raw), Smooth (Sample)). (a) Original: ranges differ across channels; (b) Smooth (Raw): align maxima only; (c) Smooth (Sample): align the main body (excluding outliers), improving suitability for quantization.}
\label{fig:smooth}
\vspace{-8.mm}
\end{figure*}

\textbf{\textit{Challenge II. Inaccurate smoothing factor estimation.}} When the maximum absolute value, which is an outlier, is used to compute the smoothing factor, the maximum values across channels can be aligned. However, as shown in Fig.~\ref{fig:smooth} (Smooth (Raw)), significant range disparities remain among the non-outlier values across channels due to the large difference between outliers and non-outliers.

Some existing PTQ methods ~\cite{li2019percentile, OmniQuant} are specifically designed to address the problems caused by outliers. However, they are not designed for the outlier distributions of demoir\'eing models. Consequently, they fail to mitigate quantization errors.

\begin{figure*}[t]
\begin{center}
\hspace{-0.mm}\includegraphics[width=1.\linewidth]{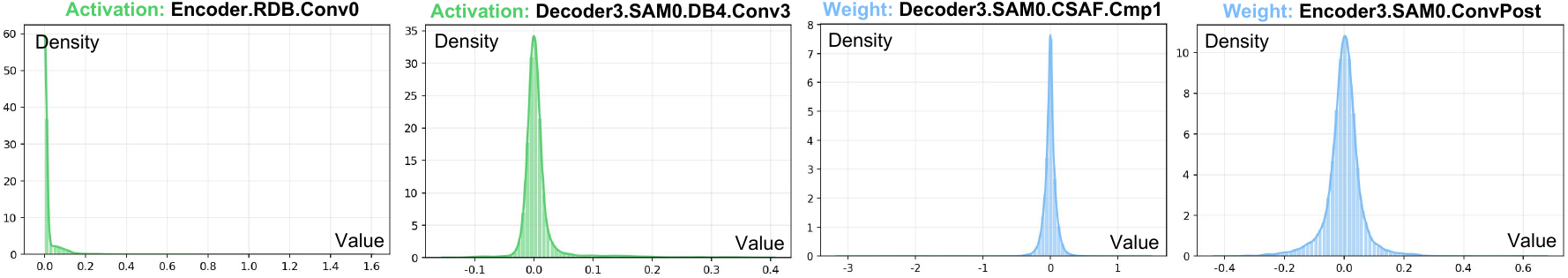} \\
\end{center}
\vspace{-1.mm}
\caption{Visualization of \textbf{activation} and \textbf{weight} distributions from randomly selected layers in ESDNet~\cite{yu2022towards}. The distributions are approximately Gaussian or exponential. Under these distributions, outliers are distributed on both sides. These sparse extremes can dominate the quantization range if they are treated together with ordinary values.}
\label{fig:distribution}
\vspace{-6.mm}
\end{figure*}

\noindent \textbf{Sampling-Based Activation Quantizer.} To address the problems caused by outliers, we propose a random sampling-based method for estimating activation ranges. From Fig.~\ref{fig:distribution},  most activations conform to either a Gaussian or an exponential distribution. Such distributions typically exhibit a small number of outliers that deviate from the normal range on one or both sides. Appendix~\ref{sec:visual_distribution} provides additional activation and weight distribution visualizations to support this point. Furthermore, we validate this distributional observation with Q-Q analysis in Appendix~\ref{sec:qq_analysis}.

To mitigate the impact of these outliers on accuracy, we adopt a strategy of randomly sampling a small subset of activation values in a channel or tensor (Fig.~\ref{fig:overview}a). $\text{Sample}\gamma(\cdot)$ denotes a subset formed by randomly selecting a proportion $\gamma$ of elements from the activations. This approach reduces the probability of selecting outliers, thereby reducing the quantization error. It also ensures an accurate estimation of the distribution range of the majority of non-outlier values. 

{In LLMs, outliers arise from the large magnitudes of certain activation channels~\cite{xiao2023smoothquant}. This allows the smoothing factor to be calculated directly from the per-channel maximum absolute value. In contrast, demoir\'eing models, which are primarily composed of convolutional blocks, feature outliers that are mostly sporadic elements. Consequently, the presence of these elemental outliers within a channel can skew the estimation of its overall magnitude. Therefore,} we estimate the maximum magnitude of activations for each channel by randomly sampling: 
\begin{equation}
\begin{gathered}
M_j = \text{max} (\lvert \text{Sample}_{\gamma_1}(\mathbf{X}_j) \rvert),
\label{eq:M_j}
\end{gathered}
\end{equation}
where $\mathbf{X}_j$ is the $j$-th input channel of activation, $M_j$ is the maximum estimated, and $\gamma_1$ is the proportion of elements sampled from each activation channel. By restricting the estimation to a sampled subset, we effectively capture the main body of activations while reducing the influence of outlier values. For the calculation of the maximum weight magnitude, we directly take the largest absolute value. The computation of the smoothing factor can be written as:
\vspace{1.mm}
\begin{equation}
\begin{gathered}
s_j = M_j^{\alpha} / \text{max} (\lvert \mathbf{W}_j \rvert)^{1-\alpha}.
\label{eq:s_j}
\end{gathered}
\end{equation}

Visualizations in Fig.~\ref{fig:smooth} reveal that calculating smoothing factor with sampled maximum magnitude of activations (Smooth (Sample)) achieves better smoothing of non-outlier values (\textit{\textit{i.e.}}, the majority of the distribution). 
This allocates more quantization levels to ordinary activations.
After being divided by the smoothing factor, the activation's per-channel magnitude have been aligned. However, the outlier elements still remain in the activation tensor. These outliers barely improve output quality but greatly increase quantization error, so they should be clipped during quantization. Appendix~\ref{sec:outlier_sensitivity} provides a sensitivity study showing that clipping a tiny fraction of activation outliers changes the convolution output by less than 0.01\%, supporting their limited contribution to reconstruction. Therefore, in computing the quantization bound, we still employ a sampling approach, with a proportion $\gamma_2$:
\begin{equation}
\begin{gathered}
u_a = \text{max}(\text{Sample}_{\gamma_2}(\mathbf{X}\oslash \mathbf{s})), \\ 
l_a = \text{min}(\text{Sample}_{\gamma_2}(\mathbf{X}\oslash \mathbf{s})),
\label{eq:ul_a}
\end{gathered}
\end{equation}
where $l_a$, $u_a$ denote the lower and upper bounds of the quantizer. $\mathbf{X}$ is the activation tensor and $\mathbf{s}$ is the smoothing factor. {Experiments show that computing the quantization bound via random sampling outperforms clipping activations at a fixed percentile in reducing quantization error for demoir\'eing models. This makes the bound adaptive to tensor size.} A mathematical analysis in Appendix~\ref{sec:app_sampling_theory} shows the equivalence between sampling and dynamic percentile clipping, and explains why sampling better matches activation outlier distributions across input sizes.

Finally, the derived $l_a$ and $u_a$ are applied in Eq.~\eqref{eq:quantize} to perform activation quantization, forming the sampling-based activation quantizer. This quantizer effectively reduces the impact of outliers on quantization accuracy, and improves model performance.

To avoid extra inference-time overhead, the outlier-aware quantizer uses static quantization. We compute the smoothing factor and quantization bounds using a calibration set of 200 training dataset pairs. The calibrated parameters are then reused for all test images. In Appendix~\ref{sec:stability_reproducibility}, we report the variance over multiple runs, showing that our stochastic sampling procedure remains stable.

\noindent \textbf{Mixed-Precision Weight Quantizer.} As illustrated in Fig.~\ref{fig:distribution}, similar to the behavior observed in activation distributions, the weight distribution of the network exhibits an approximately Gaussian distribution. Therefore, outliers of the weight exist on the left and right side of the distribution. A straightforward way to handle these extreme outlier values is to apply the same sampling-based estimation method proposed in activation quantization. However, unlike activations, weight tensors {of convolutional modules} generally contain far fewer elements. Consequently, the direct truncation of extreme weight values leads to a notable degradation in the performance of the quantized model. This indicates activation and weight outliers should be treated differently: activation outliers are suppressed through range estimation, whereas weight outliers need to preserve their numerical contribution.

One solution is to store outliers in FP16 via mixed precision. However, most existing quantization methods apply mixed precision at the layer or channel granularity~\cite{feng2025mpq}. 
Because outliers in convolutional-module weights occur sporadically at random positions, these coarse-grained schemes are not directly applicable. They would keep many normal weights in FP16 unnecessarily.

To overcome this limitation, we propose preserving the outliers in FP16 precision. The illustration is in Fig.~\ref{fig:overview}a. By preserving a proportion $\beta$ of outliers at both ends, we can calculate the corresponding lower and upper percentile thresholds: $T_{\mathrm{low}} = P_{\beta}(\mathbf{W})$ and $
T_{\mathrm{high}} = P_{1-\beta}(\mathbf{W})$. Here, 
$P_{\beta}(\mathbf{W})$ denotes the 
$\beta$-th percentile of $\mathbf{W}$. The complete weight quantization process can then be expressed:
\begin{equation}
\begin{gathered}
\mathbf{W}_{\text{outlier}} = \{ w_i \in \mathbf{W} \mid w_i < T_{\mathrm{low}} \;\; \text{or} \;\; w_i > T_{\mathrm{high}} \}, \\
\mathbf{W}_{\text{normal}} = \mathbf{W} \setminus \mathbf{W}_{\text{outlier}}, \quad \hat{\mathbf{W}} = \mathbf{W}_{\text{outlier}} \cup Q(\mathbf{W}_{\text{normal}}),
\label{eq:mixed_weight} 
\end{gathered}
\end{equation}
where $\mathbf{W}_{\text{outlier}}$ and $\mathbf{W}_{\text{normal}}$ are the outlier and non-outlier weights; $\hat{\mathbf{W}}$ is the final quantized weight.
Experimental results in Tab.~\ref{tab:ablation_weight} indicate that $\beta = 0.005$ is sufficient.

During inference, we decompose the convolution into two branches: a low-bit branch and an FP16 branch. The low-bit branch performs the convolution using quantized weights and quantized activations. The FP16 branch computes, on an element-wise basis, the contributions of stored weight outliers to the output. The results of the two branches are then summed. Thus, most computation remains low-bit. Because only a very small fraction of weight outliers are stored in FP16, under W4A4, the FP16 branch introduces only a 6\% memory and a 7\% runtime overhead.

\noindent \textbf{Conclusion.} The outlier-aware quantizer reduces errors caused by outliers in activations and weights. It uses random sampling to estimate the activation range and selectively retain weight outliers in FP16, reducing quantization errors with little overhead. This dual strategy balances efficiency with accuracy. Appendix~\ref{sec:app_pseudocode} provides pseudocode for the sampling-based activation quantizer.

\vspace{-2.mm}
\subsection{Frequency-aware calibration strategy}
\vspace{-2.mm}
\label{sec:training}

Optimizing quantization boundaries through training can improve the performance to some extent. However, the quantized model still exhibits performance degradation. In particular, quantization affects frequency components unevenly. For high-frequency structures such as edges and textures, the distortions are relatively minor due to their sharp RGB variations. However, in low- and mid-frequency regions with gradual intensity transitions, quantization severely reduces representational capacity, often manifesting as color banding. Appendix~\ref{sec:app_banding} visualizes this banding effect and the improvement brought by frequency-aware calibration. 

To address the above problem, we propose a frequency-aware calibration strategy. This method optimizes the parameters of the quantizer by extracting the mid- and low-frequency components from both the model outputs and the ground truth, then optimizing the loss accordingly. A standard spatial-domain loss can be dominated by edges and textures, making it less effective at penalizing low-amplitude errors in smooth mid- and low-frequency regions. Therefore, FAC explicitly aligns the calibration objective with the frequency bands where banding artifacts are most visible.
  Specifically, following previous work~\cite{liu2025freqformer,wang2024}, we employ a dedicated convolution kernel $\vk$ to extract the mid- and low-frequency components. The kernel is defined as:
\begin{equation}
\begin{gathered}
\vk = \begin{bmatrix}
{1/16} & {1/8} & {1/16} \\
{1/8} & {1/4} & {1/8} \\
{1/16} & {1/8} & {1/16}
\end{bmatrix}.
\label{eq:kernel_k}
\end{gathered}
\end{equation}
The frequency extraction process is an $L$-step iterative procedure based on kernel $\vk$. Given a image $\mathbf{I}$, in the 
$i$-th step ($i$ = 1, 2, ..., $L$), the extraction can be formalized as: 
\begin{equation}
\begin{gathered}
\mathbf{I}^{i} = E_{\vk}^i(\mathbf{I}^{i-1}),
\label{eq:frequency_step}
\end{gathered}
\end{equation}
where $\mathbf{I}^0$$=$$\mathbf{I}$, $\mathbf{I}^{i}$ is the outcome of the $i$-th step of extraction. $E_{\vk}^i$ denotes an operation where the input is convolved with the kernel $\vk$ and a dilation of $2^i$. The whole \textbf{frequency extraction} process can be written as a recursive function $F(\cdot)$, which is defined as follows:
\begin{equation}
\begin{gathered}
F(\mathbf{I}, i) =
\begin{cases}
\mathbf{I}, & i = 0, \\
E_{\vk}^i\!\big(F(\mathbf{I}, i-1)\big), & i \geq 1.
\end{cases}
\label{eq:extraction_function}
\end{gathered}
\end{equation}
The outcome of the whole frequency extraction procedure is $F(\mathbf{I}, L)$. As Fig.~\ref{fig:overview}b, during the phase of optimizing quantization boundaries, we extract the low- and mid-frequency components from the output image and ground truth. This keeps the calibration target in the image domain while emphasizing smooth regions. {Then following ESDNet~\cite{yu2022towards}, we directly reuse these losses: a pixel-wise $\mathcal{L}_1$ loss and a feature-based perceptual loss $\mathcal{L}_p$.} The perceptual loss $\mathcal{L}_p$ is computed as the $L_1$ distance between the VGG16 extracted feature~\cite{yu2022towards}. 

Specifically, given the output of the quantitative model $\mathbf{I}'$ and the corresponding ground truth $\hat{\mathbf{I}}$, the extracted component of low and mid frequencies are defined as $\mathbf{I}'_l = F(\mathbf{I}', L)$ and $\hat{\mathbf{I}}_l =F(\hat{\mathbf{I}}, L)$, respectively. Then, we compute the loss $\mathcal{L}_{\text{total}}$ on the extracted frequencies as follows:
\begin{equation}
\begin{gathered}
\mathcal{L}_{\text{total}} = \mathcal{L}_1(\mathbf{I}'_l, \hat{\mathbf{I}}_l) + \mathcal{L}_p(\mathbf{I}'_l, \hat{\mathbf{I}}_l).
\label{eq:loss_total}
\end{gathered}
\end{equation}
We only optimize the quantization boundaries of activations while keeping all other parameters fixed. Besides, we find that when $L$ is small (\textit{\textit{e.g.}}, $L=1$), more high-frequency information is preserved, whereas a larger $L$ (\textit{\textit{e.g.}}, $L=5$) focuses only on low frequencies. Since our goal is to focus on both low- and mid-frequency regions, we set $L=3$. Experiments in Sec.~\ref{sec:ablation} (Tab.~\ref{tab:abl_level}) validate this choice. Since the extraction module introduces only a few convolutions, the overhead is negligible. Besides, FAC changes the calibration signal rather than the model, and therefore does not affect inference.

\vspace{-1.mm}
\section{Experiments}
\vspace{-1.mm}

\subsection{Experimental Settings}
\label{sec:setting}
\noindent \textbf{Datasets and Metrics.} We evaluate on three moir\'e pattern removal datasets: UHDM~\cite{yu2022towards}, FHDMi~\cite{he2020fhde2net}, and LCDMoir\'e~\cite{yuan2019aim}. For each dataset, we randomly select 200  training image pairs and crop them to construct a calibration dataset. After the calibration stage, we evaluate the results on the corresponding test dataset. For quantitative assessment, we adopt PSNR, SSIM~\cite{wang2004image}, and LPIPS~\cite{zhang2018unreasonable}. 

\noindent \textbf{Implementation Details.} We adopt ESDNet~\cite{yu2022towards} as the backbone and provide results under 3, 4, 6, and 8 bits. The quantization setup is denoted as W$w$A$a$, where $w$ and $a$ are the bit widths of weights and activations. For the sampling rate, we set $\gamma_1 = \gamma_2 = 10^{-3}$. For the smoothing factor, we use $\alpha = 0.5$. Both our method and the compared baselines apply static quantization, with per-channel quantization for weights and per-tensor quantization for activations. The baselines and our method are fairly evaluated, with more details provided in Appendix~\ref{sec:app_comparison_methods}. For the FP16 weight branch, since there’s currently no way to handle FP16 convolution outliers, we implemented a fused CUDA kernel.

\noindent \textbf{Calibration Settings.} During the optimization stage, the quantization parameters are trained for 4 epochs using the Adam~\cite{kingma2014adam} optimizer with $\beta_1 = 0.9 $ and $\beta_2 = 0.999$. The image pairs are randomly cropped into $512 \times 512$, and the batch size is $1$. We apply the loss proposed in Eq.~\eqref{eq:loss_total}. The initial learning rate is set to $0.001$ and scheduled with cyclic cosine annealing~\cite{loshchilov2016sgdr}. All codes are implemented with PyTorch~\cite{paszke2019pytorch}, and all experiments are conducted on one A6000-48G GPU.

\begin{table*}[t]
\scriptsize
\vspace{-2.5mm}
\centering
\captionsetup[subfloat]{position=bottom}
\hspace{-2.mm}
\subfloat[Quantizer bound calculation. \vspace{-1.mm}
\label{tab:abl_bound}]{
        \scalebox{1.}{
        \setlength{\tabcolsep}{3.2mm}
            \begin{tabular}{l | c c c c c c c}
            \toprule
            \rowcolor{color3} Method & MinMax & Percentile & Sample\\
            \midrule
            PSNR $\uparrow$ & 16.51  & 16.85  & \textbf{17.61}\\
            SSIM $\uparrow$ & 0.5255  & 0.6701  & \textbf{0.6892} \\
            LPIPS $\downarrow$ & 0.6786  & 0.4639  & \textbf{0.4552}\\
            \bottomrule
            \end{tabular}
}}\hspace{1.mm}
\subfloat[Smooth transformation. \vspace{-1.mm}
\label{tab:abl_smooth}]{
        \scalebox{1.}{
        \setlength{\tabcolsep}{3.1mm}
            \begin{tabular}{l | c c c c c c c}
            \toprule
            \rowcolor{color3} Method & Original & Smooth (Raw) & Smooth (Sample) \\
            \midrule
            PSNR $\uparrow$ & 17.61 & 20.43   & \textbf{20.52}\\
            SSIM $\uparrow$ & 0.6892 & 0.7408 & \textbf{0.7538} \\
            LPIPS $\downarrow$ & 0.4552 & 0.3363   & \textbf{0.3247 }\\
            \bottomrule
            \end{tabular}
}}
\caption{Ablation study on sampling-based activation quantizer.}
\label{tab:ablation_activation}
\vspace{-1.mm}
\end{table*}

\begin{table*}[t]
\scriptsize
\hspace{-2.mm}
\centering
\subfloat[Outlier handling method. \vspace{-1.mm} 
\label{tab:abl_weight_outlier}]{
        \scalebox{1.}{
        \setlength{\tabcolsep}{1.mm}
            \begin{tabular}{l | c c c c c c c}
            \toprule
            \rowcolor{color3} Method & Baseline & Discard Outliers & Store Random & Store Outlier \\
            \midrule
            PSNR $\uparrow$ & 20.52 & 16.29 & 20.54  & \textbf{20.92} \\
            SSIM $\uparrow$ & 0.7538   & 0.6724 & 0.7542 & \textbf{0.7570} \\
            LPIPS $\downarrow$ & 0.3247  & 0.4103 & 0.3244 & \textbf{0.3171} \\
            \bottomrule
            \end{tabular}
}}\hspace{0.mm}
\subfloat[Outliers Proportion ($\beta$). \vspace{-1.mm}
\label{tab:abl_weight_ratio}]{
        \scalebox{1.}{
        \setlength{\tabcolsep}{1.1mm}
            \begin{tabular}{l | c c c c c c c}
            \toprule
            \rowcolor{color3} \#$\beta$ & PSNR $\uparrow$ & SSIM $\uparrow$ & LPIPS $\downarrow$ & Params (M) & OPs (G) \\
            \midrule
            $0.25\%$ & 20.88 & 0.7565 & 0.3180 & 0.7837 & 1.78\\ 
            $0.50\%$ & 20.92 & 0.7570 & 0.3171  & 0.7948 & 1.80 \\
            $1.00\%$ & 21.04 & 0.7578 & 0.3171 & 0.8171 & 1.85 \\
            \bottomrule
            \end{tabular}
}}
\caption{Ablation study on mixed-precision weight quantizer.} 
\vspace{-4.mm}
\label{tab:ablation_weight}
\end{table*}

\begin{table*}[t]
\scriptsize
\captionsetup[subfloat]{position=bottom}
\centering
\hspace{-1.mm}
\subfloat[Calibration strategy. \vspace{-1.mm} 
\label{tab:abl_training_method}]{
        \scalebox{1.}{
        \setlength{\tabcolsep}{2.3mm}
            \begin{tabular}{l | c c c c c c c}
            \toprule
            \rowcolor{color3}  Method & No Calibration & Orignal Image & High-Freq & Mid\&Low-Freq \\
            \midrule
            PSNR $\uparrow$ & 20.92   & 20.95 & 20.62 & \textbf{21.08}\\
            SSIM $\uparrow$ & 0.7570   & 0.7604 & 0.7538 & \textbf{0.7626} \\
            LPIPS $\downarrow$ & 0.3171  & 0.3088 & 0.3275  & \textbf{0.3068}\\
            \bottomrule
            \end{tabular}
}}\hspace{0.mm}
\subfloat[Frequency extraction level. \vspace{-1.mm} 
\label{tab:abl_level}]{
        \scalebox{1.}{
        \setlength{\tabcolsep}{2.3mm}
            \begin{tabular}{l | c c c c c c c}
            \toprule
            \rowcolor{color3} Level & 1 & 3 & 5 \\
            \midrule
            PSNR $\uparrow$ & 20.96   & \textbf{21.08} & 21.05 \\
            SSIM $\uparrow$ & 0.7605 & \textbf{0.7626} &  0.7605\\
            LPIPS $\downarrow$ & 0.3092  & \textbf{0.3068}& 0.3071 \\
            \bottomrule
            \end{tabular}
}}
\caption{Ablation study on frequency-aware calibration strategy.}
\vspace{-2.mm}
\label{tab:ablation_traning}
\end{table*}

\subsection{Ablation Study}
\label{sec:ablation}
\vspace{-1.mm}
We investigate the impact of the proposed outlier-aware quantizer (for activations and weights) and the frequency-aware calibration strategy. All experiments are conducted on the UHDM~\cite{yu2022towards} dataset with 4-bit quantization (W4A4). The settings follow Sec.~\ref{sec:setting} to ensure fairness. 

\noindent \textbf{Activation Quantizer.} 
Tab.~\ref{tab:abl_bound} compares activation bound estimation methods. The baseline is the min-max method~\cite{jacob2017fakequant}. We compare it with the percentile~\cite{li2019percentile} and our sample-based method (as in Eq.~\eqref{eq:s_j}). Compared with previous boundary selection, our sampling-based approach is more effective.

Furthermore, we introduce a smooth transformation to adjust channels within the sample-based framework. As illustrated in Fig.~\ref{fig:smooth}, we compare two variants: the original definition and our sample-based smoothing. Results in Tab.~\ref{tab:abl_smooth} show that outlier removal via sampling enhances the effectiveness of the smoothing. In general, our sample-based activation quantizer leads to more robust quantization.

\noindent \textbf{Weight Quantizer.} 
We compare different strategies for handling weight outliers in Tab.~\ref{tab:abl_weight_outlier}. We adopt the smooth (sample-based) result from Tab.~\ref{tab:abl_smooth} as the baseline. The results show that, compared with randomly storing some weights in FP16, explicitly preserving outliers in FP16 is more effective. Besides, consistent with the analysis in Sec.~\ref{sec:quantizer}, discarding outliers leads to a performance drop.

We further evaluate the proportion of outliers ($\beta$) in Tab.~\ref{tab:abl_weight_ratio}. The OPs are computed under an input size of $3$$\times$$224$$\times$$224$. We find that setting the ratio $\beta$$=$$0.5\%$ achieves a favorable trade-off between efficiency and performance. Therefore, we adopt this setting as the default in QuantDemoire.

\noindent \textbf{Calibration Strategy.} 
We conduct an ablation study on different calibration strategies for optimizing quantized parameters. We apply the outlier-aware quantizer as the baseline. Results in Tab.~\ref{tab:abl_training_method} show that focusing on mid- and low-frequency components yields better performance. 

We also evaluate frequency extraction levels (Tab.~\ref{tab:abl_level}). The relatively high (level=1) or low (level=5) frequencies are not optimal. In contrast, level=3 achieves better performance, which aligns with our analysis. 
Besides, traditional image quality assessment (IQA) metrics do not accurately reflect the impact of banding artifacts \cite{wang2016perceptual}. Therefore, Appendix~\ref{sec:app_banding} presents additional visual results comparing the original spatial-domain loss with frequency-aware calibration.
Appendix~\ref{sec:app_more_ablation} and Appendix~\ref{sec:app_ablation_details} provide additional sampling-rate ablations and the full settings for all ablation variants.

\begin{table*}[t]
\centering
\resizebox{\textwidth}{!}{
\setlength{\tabcolsep}{3.mm}
\renewcommand{\arraystretch}{0.92}
\begin{tabular}{l|c|ccc|ccc|cccc}
\hline
\toprule[0.15em]
\rowcolor{color3} &  & \multicolumn{3}{c|}{UHDM} & \multicolumn{3}{c|}{FHDMi} & \multicolumn{3}{c}{LCDMoir\'e}  \\
\rowcolor{color3} \multirow{-2}{*}{Method} & \multirow{-2}{*}{Bit} & 
PSNR$\uparrow$ & 
SSIM$\uparrow$ & 
LPIPS$\downarrow$ & 
PSNR$\uparrow$ & 
SSIM$\uparrow$ & 
LPIPS$\downarrow$ & 
PSNR$\uparrow$ & 
SSIM$\uparrow$ & 
LPIPS$\downarrow$  \\ 
\midrule[0.15em]
\multicolumn{1}{l|}{ESDNet~\cite{yu2022towards}} & \multicolumn{1}{c|}{W32A32} & 22.12  & 0.7956   & 0.2551  & 24.50   & 0.8351   & 0.1354  & 44.83   & 0.9963    & 0.0097 \\
\midrule
\multicolumn{1}{l|}{MinMax~\cite{jacob2017fakequant}} & \multicolumn{1}{c|}{W8A8} & 21.50  & 0.7727  & 0.2596  & 20.30  & 0.7631  & 0.2600  & 37.92  & 0.9865  & 0.0236  \\
\multicolumn{1}{l|}{Percentile~\cite{li2019percentile}} & \multicolumn{1}{c|}{W8A8} & 19.38  & 0.7744  & 0.2784  & 19.73  & 0.7519  & 0.2734  & 30.11  & 0.9585  & 0.0281  \\
\multicolumn{1}{l|}{2DQuant~\cite{liu20242dquant}} & \multicolumn{1}{c|}{W8A8} & 21.20  & 0.7827  & 0.2749  & 20.57  & 0.7861  & 0.2034  & \textcolor{blue}{41.60} & \textcolor{blue}{0.9923} & 0.0214 \\
\multicolumn{1}{l|}{SVDQuant~\cite{li2024svdquant}} & \multicolumn{1}{c|}{W8A8} & \textcolor{blue}{21.80}  & \textcolor{blue}{0.7907}  & \textcolor{blue}{0.2580}  & \textcolor{blue}{22.07}  & \textcolor{blue}{0.7966}  & \textcolor{blue}{0.1598} & 41.18 & 0.9915  & \textcolor{blue}{0.0165}  \\
\rowcolor{color-ours} \multicolumn{1}{l|}{QuantDemoire (ours)} & \multicolumn{1}{c|}{W8A8} & \textcolor{red}{22.00}  & \textcolor{red}{0.7932}  & \textcolor{red}{0.2555}  & \textcolor{red}{22.23}  & \textcolor{red}{0.8026}  & \textcolor{red}{0.1591}  & \textcolor{red}{42.16}  & \textcolor{red}{0.9930}  & \textcolor{red}{0.0126}  \\

\midrule

\multicolumn{1}{l|}{MinMax~\cite{jacob2017fakequant}} & \multicolumn{1}{c|}{W6A6} & 20.48  & \textcolor{blue}{0.7648}  & \textcolor{blue}{0.2828}  & 19.85  & 0.7362  & 0.3040  & 24.53  & 0.9337  & 0.1692  \\
\multicolumn{1}{l|}{Percentile~\cite{li2019percentile}} & \multicolumn{1}{c|}{W6A6} & 18.44  & 0.7562  & 0.2957  & 19.22  & 0.7353  & 0.2953  & 22.46  & 0.9342  & \textcolor{blue}{0.1074} \\
\multicolumn{1}{l|}{2DQuant~\cite{liu20242dquant}} & \multicolumn{1}{c|}{W6A6} & 20.02  & 0.7595  & 0.2893  & 20.66  & \textcolor{blue}{0.7389}  & \textcolor{blue}{0.2558} & 26.53 & 0.9366 & 0.2241 \\
\multicolumn{1}{l|}{SVDQuant~\cite{li2024svdquant}} & \multicolumn{1}{c|}{W6A6} & \textcolor{blue}{20.86}  & 0.7602  & 0.3015  & \textcolor{blue}{20.91}  & 0.6883  & 0.3732  & \textcolor{blue}{32.10} & \textcolor{blue}{0.9691}  & 0.1314  \\
\rowcolor{color-ours} \multicolumn{1}{l|}{QuantDemoire (ours)} & \multicolumn{1}{c|}{W6A6} & \textcolor{red}{21.61}  & \textcolor{red}{0.7874}  & \textcolor{red}{0.2572}  & \textcolor{red}{21.63}  & \textcolor{red}{0.7861}  & \textcolor{red}{0.1721}  & \textcolor{red}{40.48}  & \textcolor{red}{0.9910}  & \textcolor{red}{0.0206}  \\

\midrule

\multicolumn{1}{l|}{MinMax~\cite{jacob2017fakequant}} & \multicolumn{1}{c|}{W4A4} & 16.51  & 0.5255  & 0.6786  & 16.01  & 0.4958  & 0.6686  & 15.14  & 0.7031  & 0.7500  \\
\multicolumn{1}{l|}{Percentile~\cite{li2019percentile}} & \multicolumn{1}{c|}{W4A4} & 16.85  & \textcolor{blue}{0.6701}  & \textcolor{blue}{0.4639}  & \textcolor{blue}{16.92}  & \textcolor{blue}{0.6424}  & \textcolor{blue}{0.4858}  & 14.95  & \textcolor{blue}{0.8290}  & \textcolor{blue}{0.5596}  \\
\multicolumn{1}{l|}{2DQuant~\cite{liu20242dquant}} & \multicolumn{1}{c|}{W4A4} & \textcolor{blue}{17.07}  & 0.6288  & 0.5117  & 14.80  & 0.3579  & 0.7107  & 15.11  & 0.6075  & 0.7561  \\
\multicolumn{1}{l|}{SVDQuant~\cite{li2024svdquant}} & \multicolumn{1}{c|}{W4A4} & 14.68 & 0.5762  & 0.6559  & 12.43  & 0.2818  & 0.7961  & \textcolor{blue}{15.25}  & 0.7201 & 0.6687  \\
\rowcolor{color-ours} \multicolumn{1}{l|}{QuantDemoire (ours)} & \multicolumn{1}{c|}{W4A4} & \textcolor{red}{21.08}  & \textcolor{red}{0.7626}  & \textcolor{red}{0.3068}  & \textcolor{red}{20.28}  & \textcolor{red}{0.6745}  & \textcolor{red}{0.3369}  & \textcolor{red}{31.28}  & \textcolor{red}{0.9544}  & \textcolor{red}{0.1793} \\

\midrule

\multicolumn{1}{l|}{MinMax~\cite{jacob2017fakequant}} & \multicolumn{1}{c|}{W3A3} & 12.53  & 0.3802  & 0.8630  & 12.16  & 0.2903  & 0.8514 & 10.11  & \textcolor{blue}{0.6051}  & \textcolor{blue}{0.8376} \\
\multicolumn{1}{l|}{Percentile~\cite{li2019percentile}} & \multicolumn{1}{c|}{W3A3} & 14.79  & \textcolor{blue}{0.5206}  & \textcolor{blue}{0.6869}  & \textcolor{blue}{14.33}  & \textcolor{blue}{0.4444}  & \textcolor{blue}{0.7496}  & 9.30  & 0.5242 & 1.0272  \\
\multicolumn{1}{l|}{2DQuant~\cite{liu20242dquant}} & \multicolumn{1}{c|}{W3A3} & 11.20  & 0.2465  & 0.8460  & 10.92  & 0.2879  & 0.8612  & 9.07  & 0.3026  & 1.0153  \\
\multicolumn{1}{l|}{SVDQuant~\cite{li2024svdquant}} & \multicolumn{1}{c|}{W3A3} & \textcolor{blue}{14.83}  & 0.4547  & 0.7289  & 9.34  & 0.2127  & 0.8840  & \textcolor{blue}{10.17}  & 0.5658  & 0.9753  \\
\rowcolor{color-ours} \multicolumn{1}{l|}{QuantDemoire (ours)} & \multicolumn{1}{c|}{W3A3} & \textcolor{red}{19.12} & \textcolor{red}{0.6839}  & \textcolor{red}{0.4567}  & \textcolor{red}{18.26}  & \textcolor{red}{0.5722}  & \textcolor{red}{0.4930}  & \textcolor{red}{22.02}  & \textcolor{red}{0.8558}  & \textcolor{red}{0.4544}  \\
\bottomrule[0.15em]
\end{tabular}
}
\caption{Quantitative comparison. The best and second-best results are colored \textcolor{red}{red} and \textcolor{blue}{blue}.}
\vspace{-4.mm}
\label{tab:quantitative-comparison}
\end{table*}

\begin{figure*}[!t]
\scriptsize
\centering
\renewcommand{\arraystretch}{0.93}
\begin{tabular}{ccccccccc}

\hspace{-0.48cm}
\begin{adjustbox}{valign=t}
\begin{tabular}{c}
\includegraphics[width=0.2204\textwidth]{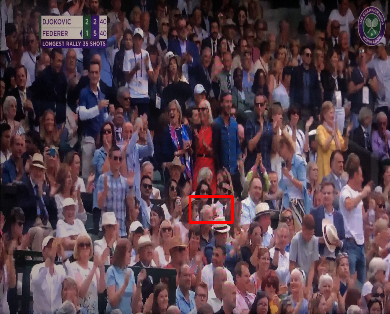}
\\
FHDMi: 00206
\end{tabular}
\end{adjustbox}
\hspace{-0.46cm}
\begin{adjustbox}{valign=t}
\begin{tabular}{cccccc}
\includegraphics[width=0.189\textwidth]{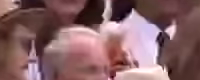} \hspace{-4.mm} &
\includegraphics[width=0.189\textwidth]{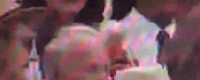} \hspace{-4.mm} &
\includegraphics[width=0.189\textwidth]{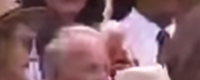} \hspace{-4.mm} &
\includegraphics[width=0.189\textwidth]{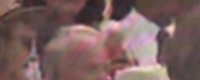} \hspace{-4.mm} &
\\ 
GT / Bit-width \hspace{-4.mm} &
Input  / - \hspace{-4.mm} &
ESDNet /  32-bit \hspace{-4.mm} &
MinMax / 6-bit \hspace{-4.mm} &
\\
\includegraphics[width=0.189\textwidth]{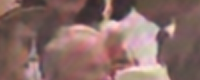} \hspace{-4.mm} &
\includegraphics[width=0.189\textwidth]{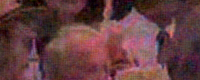} \hspace{-4.mm} &
\includegraphics[width=0.189\textwidth]{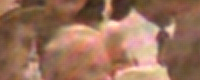} \hspace{-4.mm} &
\includegraphics[width=0.189\textwidth]{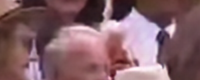} \hspace{-4.mm} &
\\ 
Percentile / 6-bit \hspace{-4.mm} &
2DQuant / 6-bit \hspace{-4.mm} &
SVDQuant / 6-bit \hspace{-4.mm} &
QuantDemoire / 6-bit \hspace{-4mm}
\\
\end{tabular}
\end{adjustbox}
\\

\hspace{-0.48cm}
\begin{adjustbox}{valign=t}
\begin{tabular}{c}
\includegraphics[width=0.2204\textwidth]{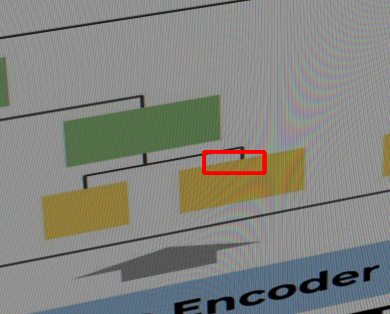}
\\
LCDMoir\'e: 000031
\end{tabular}
\end{adjustbox}
\hspace{-0.46cm}
\begin{adjustbox}{valign=t}
\begin{tabular}{cccccc}
\includegraphics[width=0.189\textwidth]{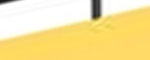} \hspace{-4.mm} &
\includegraphics[width=0.189\textwidth]{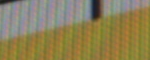} \hspace{-4.mm} &
\includegraphics[width=0.189\textwidth]{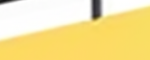} \hspace{-4.mm} &
\includegraphics[width=0.189\textwidth]{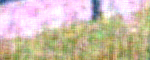} \hspace{-4.mm} &
\\ 
GT / Bit-width \hspace{-4.mm} &
Input  / - \hspace{-4.mm} &
ESDNet /  32-bit \hspace{-4.mm} &
MinMax / 4-bit \hspace{-4.mm} &
\\
\includegraphics[width=0.189\textwidth]{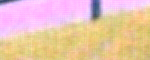} \hspace{-4.mm} &
\includegraphics[width=0.189\textwidth]{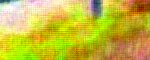} \hspace{-4.mm} &
\includegraphics[width=0.189\textwidth]{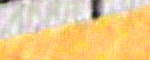} \hspace{-4.mm} &
\includegraphics[width=0.189\textwidth]{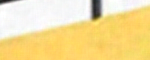} \hspace{-4.mm} &
\\ 
Percentile / 4-bit \hspace{-4.mm} &
2DQuant / 4-bit \hspace{-4.mm} &
SVDQuant / 4-bit \hspace{-4.mm} &
QuantDemoire / 4-bit \hspace{-4mm}
\\
\end{tabular}
\end{adjustbox}
\\

\hspace{-0.48cm}
\begin{adjustbox}{valign=t}
\begin{tabular}{c}
\includegraphics[width=0.2204\textwidth]{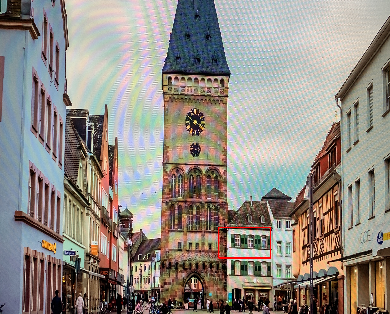}
\\
UHDM: 0013
\end{tabular}
\end{adjustbox}
\hspace{-0.46cm}
\begin{adjustbox}{valign=t}
\begin{tabular}{cccccc}
\includegraphics[width=0.189\textwidth]{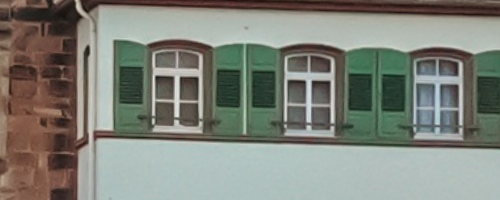} \hspace{-4.mm} &
\includegraphics[width=0.189\textwidth]{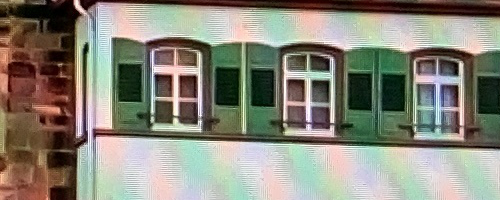} \hspace{-4.mm} &
\includegraphics[width=0.189\textwidth]{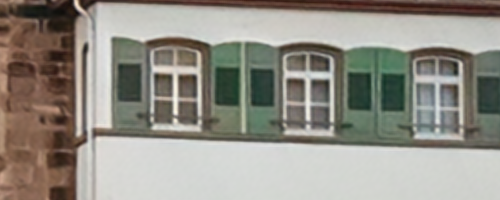} \hspace{-4.mm} &
\includegraphics[width=0.189\textwidth]{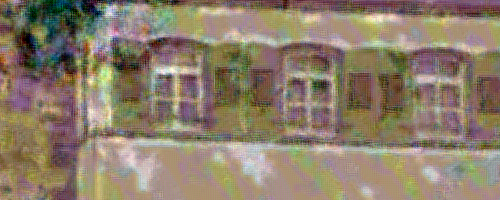} \hspace{-4.mm} &
\\ 
GT / Bit-width \hspace{-4.mm} &
Input  / - \hspace{-4.mm} &
ESDNet /  32-bit \hspace{-4.mm} &
MinMax / 3-bit \hspace{-4.mm} &
\\
\includegraphics[width=0.189\textwidth]{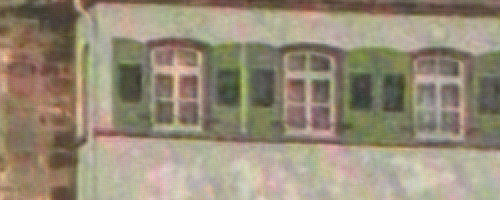} \hspace{-4.mm} &
\includegraphics[width=0.189\textwidth]{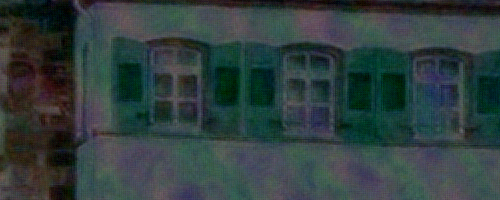} \hspace{-4.mm} &
\includegraphics[width=0.189\textwidth]{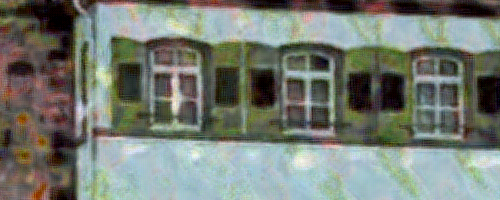} \hspace{-4.mm} &
\includegraphics[width=0.189\textwidth]{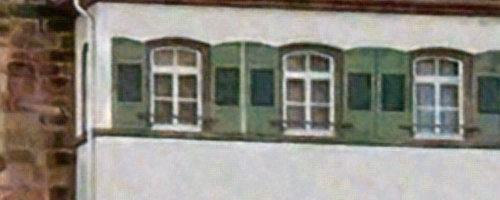} \hspace{-4.mm} &
\\ 
Percentile / 3-bit \hspace{-4.mm} &
2DQuant / 3-bit \hspace{-4.mm} &
SVDQuant / 3-bit \hspace{-4.mm} &
QuantDemoire / 3-bit \hspace{-4mm}
\\
\end{tabular}
\end{adjustbox}

\end{tabular}
\vspace{-1.mm}
\caption{Visual comparison on UHDM~\cite{yu2022towards}, FHDMi~\cite{he2020fhde2net}, and LCDMoir\'e~\cite{yuan2019aim} datasets.}
\label{fig:visual}
\vspace{-6.mm}
\end{figure*}

\vspace{-1.mm}
\subsection{Comparison with State-of-the-Art Methods}
\vspace{-1.mm}
We utilize ESDNet~\cite{yu2022towards} as the full-precision backbone and compare QuantDemoire with several advanced post-training quantization methods, including MinMax~\cite{jacob2017fakequant}, Percentile~\cite{li2019percentile}, 2DQuant~\cite{liu20242dquant}, and SVDQuant~\cite{li2024svdquant}. Moreover, Appendix~\ref{sec:app_svd_variants} further analyzes SVDQuant with different ranks, Appendix~\ref{sec:generalization} evaluates QuantDemoire on another demoir\'eing backbone and an image super-resolution model, and Appendix~\ref{sec:efficiency_analysis} reports end-to-end latency.

\noindent \textbf{Quantitative Results.} 
Quantitative evaluations are summarized in Tab.~\ref{tab:quantitative-comparison}. Across all datasets, bit-widths, and metrics, our method outperforms prior quantization methods, with larger gains at low bit-widths (\textit{e.g.}, 3/4-bit). For example, in the 4-bit setting, our approach surpasses the latest method, SVDQuant~\cite{li2024svdquant}, by more than \textbf{6 dB} on the UHDM~\cite{yu2022towards} dataset. Furthermore, our proposed method retains performance close to the full-precision network under 8/6-bit settings, whereas the degradation at 3/4-bit is greatly reduced.
These results demonstrate the effectiveness of our approach.

\noindent \textbf{Qualitative Results.} 
Fig.~\ref{fig:visual} shows 3/4/6-bit comparisons. Existing quantization baselines tend to leave residual moir\'e artifacts or fail to reconstruct clean structures, especially on difficult samples and at lower precision. Our QuantDemoire removes moir\'e patterns and restores fine details, producing results close to full precision. This observation is consistent with the quantitative results in Tab.~\ref{tab:quantitative-comparison}. Appendix~\ref{sec:app_more_qualitative} provides additional visual comparisons across datasets and bit-widths.

\begin{wraptable}{r}{0.58\textwidth}
\scriptsize
\vspace{-6.mm}
\begin{center}
\resizebox{\linewidth}{!}{
\setlength{\tabcolsep}{2.2mm}
\begin{tabular}{l|c|c|c|ccccc}
    \toprule
    \rowcolor{color3}  & Bit & Params (M) & OPs (G) & \multicolumn{2}{c}{UHDM} \\
    \rowcolor{color3}  \multirow{-2}{*}{Method} & ($w$/$a$) & ($\downarrow$Ratio) & ($\downarrow$Ratio) & PSNR $\uparrow$ & SSIM $\uparrow$ \\
    \midrule
    ESDNet & 32/32 & 5.93 ($\downarrow$0\%) & 13.52 ($\downarrow$0\%) & 22.12 & 0.7956 \\
    \midrule
    2DQuant & 6/6 & 1.12 ($\downarrow$81.14\%) & 2.54 ($\downarrow$81.25\%) & 20.02 & 0.7595\\
    SVDQuant & 6/6 & 1.45 ($\downarrow$75.53\%) & 4.60 ($\downarrow$65.97\%) & 20.86 & 0.7602\\
    \rowcolor{color-ours} QuantDemoire & 6/6 & 1.16 ($\downarrow$80.43\%) & 2.64 ($\downarrow$80.49\%) & 21.61 & 0.7874\\
    \midrule
    2DQuant & 4/4 & 0.75 ($\downarrow$87.38\%) & 1.69 ($\downarrow$87.50\%) & 17.07 & 0.6288\\
    SVDQuant & 4/4 & 1.08 ($\downarrow$81.78\%) & 3.76 ($\downarrow$72.19\%) & 14.68 & 0.5762\\
    \rowcolor{color-ours} QuantDemoire & 4/4 & 0.79 ($\downarrow$86.61\%) & 1.80 ($\downarrow$86.68\%) & 21.08 & 0.7626\\
    \midrule
    2DQuant & 3/3 & 0.56 ($\downarrow$90.51\%) & 1.27 ($\downarrow$90.63\%) & 11.20 & 0.2465\\
    SVDQuant & 3/3 & 0.90 ($\downarrow$84.89\%) & 3.34 ($\downarrow$75.30\%) & 14.83 & 0.4547\\
    \rowcolor{color-ours} QuantDemoire & 3/3 & 0.61 ($\downarrow$89.69\%) & 1.38 ($\downarrow$89.78\%) & 19.12 & 0.6839\\
    
    \bottomrule
\end{tabular}
}
\vspace{-1.mm}
\caption{Compression ratios of Params and OPs at 6/4/3-bit. OPs are measured with an input size of $3$$\times$$224$$\times$$224$.}
\vspace{-4.mm}
\label{tab:copmression}
\end{center}
\end{wraptable}

\noindent \textbf{Compression Ratio.}
We report the parameter (Params) and operation (OPs) compression ratios at 6/4/3-bit in Tab.~\ref{tab:copmression}. The OPs are measured with an input size of $3$$\times$$224$$\times$$224$. Compared with the full-precision (32-bit) model, our approach significantly reduces both model size and computational cost. At 4-bit, the compression rate exceeds 86.6\%, while the performance drop is limited to 4.7\% (PSNR). Moreover, compared with existing methods, QuantDemoire demonstrates clear advantages. Appendix~\ref{sec:app_more_compression} provides a more comprehensive compression-ratio comparison.

\vspace{-2.mm}
\section{Conclusion}
\vspace{-2.mm}
In this paper, we propose QuantDemoire, a post-training quantization method tailored for image demoir\'eing. Our design is developed from two perspectives: the quantizer and the calibration strategy. We first introduce the outlier-aware quantizer to reduce quantization errors caused by outliers. We 
Besides, we develop a frequency-aware calibration strategy emphasizing mid- and low-frequency components to mitigate banding artifacts caused by low-bit quantization. 
Comprehensive experiments verify that our method effectively reduces overhead compared with the full-precision model. Meanwhile, QuantDemoire outperforms current advanced quantization methods.

\begin{ack}
This work was supported by Huawei Explore X funds. This work was also sponsored by AI Hundred Schools Program and was carried out using the Ascend AI technology stack.
\paragraph{Competing interests.}
The authors declare no additional competing interests beyond the affiliations and funding disclosed above.
\end{ack}

{\small
\bibliographystyle{plain}
\bibliography{example_paper}
}


\newpage
\appendix

\section{More Ablation}
\label{sec:app_more_ablation}

\begin{table*}[htbp]
\scriptsize
\captionsetup[subfloat]{position=bottom}
\label{tab:ablation_rate}
\centering
\hspace{-2.mm}
\subfloat[Smooth sampling rate ($\gamma_1$). 
\label{tab:abl_bound_rate}]{
        \scalebox{1.}{
        \setlength{\tabcolsep}{4.7mm}
            \begin{tabular}{l | c c c c c c c}
            \toprule
            \rowcolor{color3}Rate & $10^{-2}$ & $10^{-3}$ & $10^{-4}$ \\
            \midrule
            PSNR $\uparrow$ & 19.92   & \textbf{20.92 } &  20.80 \\
            SSIM $\uparrow$ & 0.6941    & \textbf{0.7570 } & 0.7505  \\
            LPIPS $\downarrow$ & 0.4141     & \textbf{0.3171} & 0.3221  \\
            \bottomrule
            \end{tabular}
}}\hspace{0.mm}
\subfloat[Bound sampling rate ($\gamma_2$). 
\label{tab:abl_smooth_rate}]{
        \scalebox{1.}{
        \setlength{\tabcolsep}{4.7mm}
            \begin{tabular}{l | c c c c c c c}
            \toprule
            \rowcolor{color3}  Rate & $10^{-2}$ & $10^{-3}$ & $10^{-4}$ \\
            \midrule
            PSNR $\uparrow$ & 20.88   & \textbf{20.92 } &  20.89 \\
            SSIM $\uparrow$ & 0.7558    & \textbf{0.7570 } & 0.7474  \\
            LPIPS $\downarrow$ & 0.3183     & \textbf{0.3171} & 0.3388  \\
            \bottomrule
            \end{tabular}
}}
\caption{Ablation on the sampling rate in the activation. Evaluation is conducted on UDHM, 4-bit.}
\end{table*}

For the smooth stage ($\gamma_1$), we ablate the sampling rate by varying $\gamma_1$ while fixing $\gamma_2=10^{-3}$. As shown in Tab.~\ref{tab:abl_bound_rate}, the setting that produces the best performance corresponds to $\gamma_1=10^{-3}$. Higher rates introduce excessive outliers in the smoothing process, whereas lower rates bias the preliminary estimate. A moderate sampling rate, therefore, provides the most stable and accurate range estimation.

For the bound stage ($\gamma_2$), we very $\gamma_2$ while fixing $\gamma_1=10^{-3}$. Tab.~\ref{tab:abl_smooth_rate} indicates that $\gamma_2=10^{-3}$ gives the best results. Larger rates lead to the accumulation of outliers during refinement, and smaller rates cause biased bounds. These results confirm that a moderate sampling rate also optimizes accuracy, reinforcing our overall design choice.

\section{Details of Comparison Methods} 
\label{sec:app_comparison_methods}
\noindent \textbf{MinMax.} We apply per-channel quantization to the weights and per-tensor quantization to the activations. During the calibration, we directly use the minimum and maximum activation values as the quantization boundaries.

\noindent \textbf{Percentile.} We also use per-channel quantization for the weights and per-tensor quantization for the activations. Specifically, for both weights and activation values, the 99.9\%-th and 0.1\%-th percentiles of all values are consistently used as the quantization boundaries.

\noindent \textbf{2DQuant.} We first employ the DOBI~\cite{liu20242dquant} method to search for quantizer parameters. Then we train the model using the quantization parameters obtained from DQC~\cite{liu20242dquant}, with a training configuration of $100$ epochs, a batch size of $4$, and a learning rate of $10^{-2}$
. All other settings remain consistent with  our proposed method. 

\noindent \textbf{SVDQuant.} We first apply SmoothQuant~\cite{xiao2023smoothquant} to reduce activations outliers. Next, we perform SVD on the weights and compute both the low-rank and residual branches. To ensure fairness in our experiments and to prevent the low-rank branch from introducing an excessive number of additional parameters, we set the rank to 2.

\vspace{-1.mm}
\section{Variants of SVDQuant} 
\label{sec:app_svd_variants}
\vspace{-1.mm}

\begin{table*}[t]
\scriptsize
\begin{center}
\setlength{\tabcolsep}{4.45mm}
\begin{tabular}{l|c|c|c|ccccc}
    \toprule
    \rowcolor{color3}  &  & &  & \multicolumn{3}{c}{UHDM} \\
    \rowcolor{color3}  \multirow{-2}{*}{Method} & \multirow{-2}{*}{Bit($w$/$a$)} & \multirow{-2}{*}{Params (M)}  & \multirow{-2}{*}{OPs (G)} & PSNR $\uparrow$ & SSIM $\uparrow$ & LPIPS $\downarrow$ \\
    \midrule
    SVDQuant (rank = 2)  & 4/4 & 1.08  & 3.76 & 14.68 & 0.5762 & 0.6559\\
    SVDQuant (rank = 8) & 4/4 & 2.10 & 8.48  & 18.69 & 0.7428 & 0.3615\\
    QuantDemoire (ours) & 4/4 & \textbf{0.79} & \textbf{1.80}  & \textbf{21.08}  & \textbf{0.7626} & \textbf{0.3068}\\
    \midrule
    SVDQuant (rank = 2) & 3/3 & 0.90   & 3.34  & 14.83 & 0.4547 & 0.7289\\
    SVDQuant (rank = 8) & 3/3 & 2.01  & 6.80  & 17.20  & 0.6707 & 0.5033 \\
    QuantDemoire (ours) & 3/3 & \textbf{0.61} & \textbf{1.38}   & \textbf{19.12} & \textbf{0.6839} & \textbf{0.4567}\\
    \bottomrule
\end{tabular}
\caption{Results of SVDQuant (rank = 2 and 8). Evaluation is conducted on UDHM, 3 and 4-bit.}
\label{tab:svd_rank}
\end{center}
\end{table*}

We observed that SVDQuant suffers from a substantial performance drop when quantized to extremely low bit-widths (3 and 4 bits). To investigate this issue, we conducted further experiments with higher ranks. The results in Tab.~\ref{tab:svd_rank} show that when the rank is increased to 8, SVDQuant achieves a significant performance improvement. However, when the rank is set to 8, although the performance improves, the relatively small parameter size of the convolutional layers in ESDNet leads SVDQuant to introduce an excessive number of additional parameters. Therefore, in our experiments, we set rank to 2 for comparative fairness.

\section{More Compression Ratio} 
\label{sec:app_more_compression}
\begin{table*}[t]
\scriptsize
\begin{center}
\resizebox{\linewidth}{!}{
\setlength{\tabcolsep}{4.mm}
\begin{tabular}{l|c|c|c|ccccc}
    \toprule
    \rowcolor{color3}  & Bit & Params (M) & OPs (G) & \multicolumn{3}{c}{UHDM} \\
    \rowcolor{color3}  \multirow{-2}{*}{Method} & ($w$/$a$) & ($\downarrow$Ratio) & ($\downarrow$Ratio) & PSNR $\uparrow$ & SSIM $\uparrow$ & LPIPS $\downarrow$ \\
    \midrule
    ESDNet~\cite{yu2022towards} & 32/32 & 5.93 ($\downarrow$0\%) & 13.52 ($\downarrow$0\%) & 22.12 & 0.7956 & 0.2551\\
    \midrule
    MinMax~\cite{jacob2017fakequant} & 8/8 & 1.49 ($\downarrow$74.90\%) & 3.38 ($\downarrow$75.00\%) & 21.50 & 0.7727 & 0.2596\\
    Percentile~\cite{li2019percentile} & 8/8 & 1.49 ($\downarrow$74.90\%) & 3.38 ($\downarrow$75.00\%) & 19.38 & 0.7744 & 0.2784\\
    2DQuant~\cite{liu20242dquant} & 8/8 & 1.49 ($\downarrow$74.90\%) & 3.38 ($\downarrow$75.00\%) & 21.20 & 0.7827  & 0.2749\\
    SVDQuant~\cite{li2024svdquant} & 8/8 & 1.82 ($\downarrow$69.29\%) & 5.44 ($\downarrow$59.75\%) & 21.80 & 0.7907 & 0.2580 \\
    \rowcolor{color-ours} QuantDemoire (ours) & 8/8 & 1.53 ($\downarrow$74.25\%) & 3.47 ($\downarrow$74.31\%) & 22.00 & 0.7932 & 0.2555\\
    \midrule
    MinMax~\cite{jacob2017fakequant} & 6/6 & 1.12 ($\downarrow$81.14\%) & 2.54 ($\downarrow$81.25\%) & 20.48 &  0.7648 & 0.2828\\
    Percentile~\cite{li2019percentile} & 6/6 & 1.12 ($\downarrow$81.14\%) & 2.54 ($\downarrow$81.25\%) & 18.44 & 0.7562 & 0.2957\\
    2DQuant~\cite{liu20242dquant} & 6/6 & 1.12 ($\downarrow$81.14\%) & 2.54 ($\downarrow$81.25\%) & 20.02 & 0.7595 & 0.2893\\
    SVDQuant~\cite{li2024svdquant} & 6/6 & 1.45 ($\downarrow$75.53\%) & 4.60 ($\downarrow$65.97\%) & 20.86 & 0.7602 & 0.3015 \\
    \rowcolor{color-ours} QuantDemoire (ours) & 6/6 & 1.16 ($\downarrow$80.43\%) & 2.64 ($\downarrow$80.49\%) & 21.61 & 0.7874 & 0.2572\\
    \midrule
    MinMax~\cite{jacob2017fakequant} & 4/4 & 0.75 ($\downarrow$87.38\%) & 1.69 ($\downarrow$87.50\%) & 16.51 & 0.5255 & 0.6786\\
    Percentile~\cite{li2019percentile} & 4/4 & 0.75 ($\downarrow$87.38\%) & 1.69 ($\downarrow$87.50\%) & 16.85 & 0.6701 & 0.4639\\
    2DQuant~\cite{liu20242dquant} & 4/4 & 0.75 ($\downarrow$87.38\%) & 1.69 ($\downarrow$87.50\%) & 17.07 & 0.6288 & 0.5117 \\
    SVDQuant~\cite{li2024svdquant} & 4/4 & 1.08 ($\downarrow$81.78\%) & 3.76 ($\downarrow$72.19\%) & 14.68 & 0.5762 & 0.6559\\
    \rowcolor{color-ours} QuantDemoire (ours) & 4/4 & 0.79 ($\downarrow$86.61\%) & 1.80 ($\downarrow$86.68\%) & 21.08 & 0.7626 & 0.3068\\
    \midrule
    MinMax~\cite{jacob2017fakequant} & 3/3 & 0.56 ($\downarrow$90.51\%) & 1.27 ($\downarrow$90.63\%) & 12.53 & 0.3802 & 0.8630\\
    Percentile~\cite{li2019percentile} & 3/3 & 0.56 ($\downarrow$90.51\%) & 1.27 ($\downarrow$90.63\%) & 14.79 & 0.5206 & 0.6869\\
    2DQuant~\cite{liu20242dquant} & 3/3 & 0.56 ($\downarrow$90.51\%) & 1.27 ($\downarrow$90.63\%) & 11.20 & 0.2465 & 0.8460\\
    SVDQuant~\cite{li2024svdquant} & 3/3 & 0.90 ($\downarrow$84.89\%) & 3.34 ($\downarrow$75.30\%) & 14.83 & 0.4547 & 0.7289\\
    \rowcolor{color-ours} QuantDemoire (ours) & 3/3 & 0.61 ($\downarrow$89.69\%) & 1.38 ($\downarrow$89.78\%) & 19.12 & 0.6839 & 0.4567\\
    
    \bottomrule
\end{tabular}
}
\end{center}
\caption{Compression ratios of Params and OPs. OPs are measured with a $3$$\times$$224$$\times$$224$ input.}
\label{tab:more-copmression}
\end{table*}

A more comprehensive comparison of compression ratios is in Tab.~\ref{tab:more-copmression}. Our method achieves performance close to FP32 model, while maintaining a high compression ratio.

\section{More Qualitative Results}
\label{sec:app_more_qualitative}
Additional visual comparisons are presented in Figs.~\ref{fig:more-visual-1} and~\ref{fig:more-visual-2}. Our method demonstrates clear advantages under the 3-, 4-, and 6-bit settings, as well as across all datasets.

\section{More Distribution Visualizations}
\label{sec:visual_distribution}
Additional distributions of weights and activations are presented in Figs.~\ref{fig:act_distribution1},~\ref{fig:act_distribution2},~\ref{fig:weight_distribution1}, and~\ref{fig:weight_distribution2}. The activations in most convolutional layers approximately follow either an exponential or a Gaussian distribution, while the weights in the majority of convolutional layers exhibit an approximately Gaussian distribution overall. This further proves our point.

\subsection{{Q-Q Analysis of Activation Distributions}}
\label{sec:qq_analysis}
To further validate the distributional observation, we perform a Gaussian Q-Q analysis on the activation tensor of the layer \texttt{decoder.decoder\_2.sam\_blocks.0.fusion.compress1}. The Q-Q correlation reaches 0.9916, indicating that the main body of the activation distribution is close to Gaussian. Importantly, QuantDemoire does not depend on an exact Gaussian assumption; the key property used by our sampling estimator is the sparse occurrence of extreme outliers.

\section{Pseudocode of QuantDemoire}
\label{sec:app_pseudocode}
To provide a clearer description of the procedure of our method, we present the following pseudocode in Alg.~\ref{alg:algorithm} for the sampling-based activation quantizer.

\begin{algorithm}[t]
\caption{Pipeline of Sampling-Based Activation Quantizer}
\label{alg:algorithm}
\begin{algorithmic}
\STATE \textbf{Initialization:}
\STATE $\mathit{model} \gets \mathrm{ESDNet}(\mathrm{fp32})$
\STATE $\mathit{N} \gets$ number of batches in calibration dataset

\FOR{convolution layers $L_i$ in $\mathit{model}$}
  \STATE $\mathit{W}_i \gets$ pretrained weight of $L_i$
  \STATE $C_i \gets$ number of input channels of $L_i$
  \STATE $lb_a^i \gets 0,\; ub_a^i \gets 0$
  \STATE $lb_w^i \gets 0,\; ub_w^i \gets 0$
  \STATE $\vs^i[1:C_i] \gets 0$
\ENDFOR

\STATE \textbf{Calibration Stage 1:}
\FOR{$x$ in calibration dataset}
  \STATE $\mathit{model}(x)$
  \FOR{convolution layers $L_i$ in $\mathit{model}$}
    \STATE $\mathbf{A} \gets$ latest activation tensor of $L_i$
    \STATE $\vs^i[c] \gets \vs^i[c] + \max \big|\mathrm{Sample}_{\gamma}(\mathbf{A}_c)\big|,\ \forall c = 1,\dots,C_i$
  \ENDFOR
\ENDFOR
\FOR{convolution layers $L_i$ in $\mathit{model}$}
  \STATE $\vs^i \gets \vs^i / N$
\ENDFOR

\STATE \textbf{Calibration Stage 2:}
\FOR{$x$ in calibration dataset}
  \STATE $\mathit{model}(x)$
  \FOR{convolution layers $L_i$ in $\mathit{model}$}
    \STATE $\mathbf{A} \gets$ latest activation tensor of $L_i$
    \STATE $lb_a^i \gets lb_a^i + \min (\mathrm{Sample}_{\gamma}(\mathbf{A} \oslash \vs^i))$
    \STATE $ub_a^i \gets ub_a^i + \max (\mathrm{Sample}_{\gamma}(\mathbf{A} \oslash \vs^i))$
  \ENDFOR
\ENDFOR

\FOR{convolution layers $L_i$ in $\mathit{model}$}
  \STATE $lb_a^i \gets lb_a^i/N$
  \STATE $ub_a^i \gets ub_a^i/N$
  \STATE $W_i \gets W_i \oslash \vs^i$
\ENDFOR
\end{algorithmic}
\end{algorithm}

\section{More Details}
\label{sec:app_more_details}

\subsection{Theoretical Analysis of Sampling Approach}
\label{sec:app_sampling_theory}

Based on the following analysis, we will demonstrate that: (1) under the activation distribution of the demoir\'eing model, the sampling-based estimation is equivalent to percentile clipping with a dynamically varying percentile that adapts to the number of activations, and this dynamic adjustment better aligns with the actual distribution characteristics of outliers; (2) the variance of the equivalent percentile is small, thereby demonstrating the stability of our method.

\subsubsection{Equivalence Between Sampling and Dynamic Percentile Clipping}

As illustrated in Fig.~\ref{fig:act_distribution1} and Fig.~\ref{fig:act_distribution2}, the activation distributions of nearly all convolutional layers approximate a normal distribution following the smoothing operation.  Consequently, we model elements $N$ within a given activation tensor as $N$ independent and identically distributed random variables (i.i.d.) drawn from a standard normal distribution.  We used a sampling ratio of $\gamma = 0.1\%$, resulting in a sample size of $n = \gamma N$. Formally, let $X_1, \ldots, X_N \overset{\text{i.i.d.}}{\sim} \mathcal{N}(0,1)$ and denote the maximum of all values by $M_N = \max_{1 \leq i \leq N} X_i$. Its probability density function (PDF): 

\begin{equation}
\begin{gathered}
f_{M_N}(x) = N\phi(x)\Phi(x)^{N-1},
\label{eq:PDF}
\end{gathered}
\end{equation}

where $\phi$ and $\Phi$ are the standard normal PDF and cumulative distribution function (CDF). The expectation is then: 

\begin{equation}
\begin{gathered}
\mathbb{E}[M_N] = \int_{-\infty}^{\infty} xN\phi(x)\Phi(x)^{N-1}dx.
\label{eq:EM_N}
\end{gathered}
\end{equation}

Similarly, we can compute the expected     maximum among $n$ activation values selected at random as follows:

\begin{equation}
\begin{gathered}
\mathbb{E}[M_n] = \int_{-\infty}^{\infty} xn\phi(x)\Phi(x)^{n-1}dx.
\label{eq:EM_n}
\end{gathered}
\end{equation}

This allows us to treat the sampling-based estimation as equivalent to clipping at a percentile threshold, $p$, for our sampling-based estimation using the CDF: $p = \Phi(\mathbb{E}[M_n])$. We observe that as the total number of elements ($N$) increases, the sampling-based estimation corresponds to a higher equivalent percentile threshold.  For example, for $N=10^6$, $p \approx 99.95\%$; for $N=10^7$, $p \approx 99.995\%$; and for $N=10^8$, $p \approx 99.9995\%$.  Furthermore, we find that due to the smoothing operation, the number of extreme outliers tends to stabilize for very large $N$ ($N > 10^6$) rather than growing linearly, as shown in Tab ~\ref{tab:extreme_outlier}:

\begin{table}[t]
\scriptsize
\begin{center}
\setlength{\tabcolsep}{13.6mm}
\begin{tabular}{l|c|c}
    \toprule
    \rowcolor{color3}  {N} & {$|x-mean| > 4 * std$} & {$|x-mean| > 4.5 * std$} \\
    \midrule
    $10^6$  & 51 & 17\\
    $10^7$  & 103 & 41\\
    $10^8$  & 253 & 72\\
    \bottomrule
\end{tabular}
\end{center}
\caption{Results of number of extreme outliers  $|x-mean| > 4 * std$ means the number of activations that lies more than \textbf{four} standard deviations away from the mean.}
\label{tab:extreme_outlier}
\vspace{-4.mm}
\end{table}

Therefore, by dynamically clipping at a higher percentile for larger $N$, our method effectively removes extreme outliers while avoiding incorrectly truncate normal values.

\subsubsection{Variance Analysis}
Based on this analysis, we calculate the standard deviation of  $p$. Using a first-order Taylor expansion (the Delta Method), the standard deviation can be approximated as:

\begin{equation}
\begin{gathered}
\mathbb{E}[M_n^2] = \int_{-\infty}^{\infty} x^2 f_{M_n}(x) dx \\ = \int_{-\infty}^{\infty} x^2 n \phi(x) \Phi(x)^{n-1} dx,
\label{eq:delta_method}
\end{gathered}
\end{equation}

Numerical calculations for $N = 10^6,10^7$and  $10^8$ yield standard deviations of $6.522 \times 10^{-4}$, $6.596 \times 10^{-5}$, and $6.653 \times 10^{-6}$, respectively, which are  negligible.

\subsection{Additional Experimental Results}

\subsubsection{Stability and Reproducibility}
\label{sec:stability_reproducibility}
To validate the stability of our method, we assessed the variance by conducting five runs as shown in Tab.~\ref{tab:five_runs}.

\begin{table}[t]
\scriptsize
\begin{center}
\setlength{\tabcolsep}{13.6mm}
\begin{tabular}{c|c|c|c}
    \toprule
    \rowcolor{color3}  {RUN} & {PSNR $\uparrow$} & {SSIM $\uparrow$} & {LPIPS $\downarrow$} \\
    \midrule
    1  & 21.0808 & 0.7626 & 0.3068\\
    2  & 21.0835 & 0.7623 & 0.3066\\
    3  & 21.0798 & 0.7628 & 0.3070\\
    4  & 21.0801 & 0.7624 & 0.3059\\
    5  & 21.0824 & 0.7631 & 0.3062\\
    \midrule
    std  & 0.0014 & 0.0003 & 0.0004\\
    \bottomrule
\end{tabular}
\end{center}
\caption{Results of PSNR, SSIM, and LPIPS metrics across 5 runs.}
\label{tab:five_runs}
\vspace{-4.mm}
\end{table}

The standard deviation of each metric across five runs is negligible, demonstrating the stability of our method.

\subsubsection{More Comparison with fixed percentile}

We present additional experiments demonstrating that our sampling method outperforms fixed‑percentile clipping. Specifically, we compare sampling with 95th‑ and 99.9th‑percentile clipping.  The results in Tab.~\ref{tab:comparison_percentile} show that sampling provides clear advantages over fixed‑percentile.

\subsubsection{Sensitivity to Activation Outlier Clipping}
\label{sec:outlier_sensitivity}
We evaluate the effect of clipping rare activation outliers by comparing the convolution outputs before and after clipping. Even when clipping activations at the 99.9th percentile, the relative output error remains below 0.01\%, suggesting that these extreme activation values contribute little to reconstruction but substantially enlarge the quantization range.

\begin{table}[t]
\scriptsize
\begin{center}
\setlength{\tabcolsep}{12.6mm}
\begin{tabular}{l|c|c|c}
    \toprule
    \rowcolor{color3}  {Method} & {PSNR} & {SSIM} & {LPIPS} \\
    \midrule
    Ours (sampling)  & 20.2803 & 0.6745 & 0.3369\\
    percentile 95\%  & 19.3058 & 0.6026 & 0.3623\\
    percentile 99.9\%  & 19.3526 & 0.6091 & 0.3487\\
    \bottomrule
\end{tabular}
\end{center}
\caption{Comparison of different methods. Apart from using sampling-based estimation, \% clipping, and 99.9\% clipping to compute the quantization boundaries and the smoothing factor. The remaining settings for all three methods are identical to the final configuration of our method.}
\label{tab:comparison_percentile}
\vspace{-4.mm}
\end{table}

\subsubsection{Initialization vs. Optimization}

Because rounding in quantization can destabilize training, training alone can only reduce quantization error within a small range; a well-chosen initialization of the quantization boundary remains crucial. The sampled-activation quantizer approach aims to initialize the quantization boundary to a favorable starting point, whereas boundary optimization subsequently performs fine-grained refinement.

\begin{table}[t]
\scriptsize
\begin{center}
\setlength{\tabcolsep}{12.6mm}
\begin{tabular}{l|c|c|c}
    \toprule
    \rowcolor{color3}  {Method} & {PSNR} & {SSIM} & {LPIPS} \\
    \midrule
    Ours (sampling)  & 20.2803 & 0.6745 & 0.3369\\
    percentile 95\%  & 19.4556 & 0.6223 & 0.3601\\
    percentile 99.9\%  & 19.5781 & 0.6391 & 0.3532\\
    MinMax  & 18.9265 & 0.5921 & 0.4013\\
    \bottomrule
\end{tabular}
\end{center}
\caption{Comparison of different initialization approaches.}
\vspace{-6.mm}
\label{tab:init_opt}
\end{table}

The experiments in Tab.~\ref{tab:init_opt} substantiate this claim. We initialize the quantization boundaries using different methods, keep all other settings the same, and train each method for a sufficient number of epochs. The results show that initializing the quantization boundaries with sampling-based estimation offers a clear advantage.

This two-stage paradigm—good initialization followed by fine-grained refinement—is already established in the literature, e.g., in 2DQuant~\cite{liu20242dquant}.

\begin{table}[ht]
\scriptsize
\centering
\setlength{\tabcolsep}{7.2mm}
\begin{tabular}{l|l|l|c|c}
\toprule
\rowcolor{color3} Task / Backbone & Dataset & Method/Bit & PSNR $\uparrow$ & SSIM $\uparrow$ \\
\midrule
Demoir\'eing / FHDe$^2$Net & UHDM & MinMax/4 & 16.42 & 0.6535 \\
Demoir\'eing / FHDe$^2$Net & UHDM & Percentile/4 & 17.86 & 0.7012 \\
Demoir\'eing / FHDe$^2$Net & UHDM & QuantDemoire/4 & \textbf{20.23} & \textbf{0.7194} \\
\midrule
SR / SwinIR & Urban100 & MinMax/4 & 18.54 & 0.3871 \\
SR / SwinIR & Urban100 & Percentile/4 & 21.01 & 0.5179 \\
SR / SwinIR & Urban100 & QuantDemoire/4 & \textbf{24.15} & \textbf{0.7324} \\
\midrule
SR / SwinIR & Urban100 & MinMax/8 & 21.79 & 0.7292 \\
SR / SwinIR & Urban100 & Percentile/8 & 21.16 & 0.7205 \\
SR / SwinIR & Urban100 & QuantDemoire/8 & \textbf{25.96} & \textbf{0.8271} \\
SR / SwinIR & Urban100 & FP32/32 & 26.15 & 0.8306 \\
\bottomrule
\end{tabular}
\vspace{1.mm}
\caption{Generalization results on another demoir\'eing backbone and image super-resolution.}
\vspace{-6.mm}
\label{tab:generalization}
\end{table}

\subsubsection{Generalization to Other Backbones and Tasks}
\label{sec:generalization}
To evaluate whether QuantDemoire generalizes beyond ESDNet, we additionally test FHDe$^2$Net on UHDM under W4A4 quantization. We also evaluate SwinIR on Urban100 for image super-resolution under W4A4 and W8A8 settings. As shown in Tab.~\ref{tab:generalization}, QuantDemoire consistently outperforms MinMax and Percentile quantization across both another demoir\'eing backbone and a Transformer-based super-resolution model.

\subsection{Banding Artifacts Visualization}
\label{sec:app_banding}

Traditional image quality assessment (IQA) metrics do not accurately reflect the impact of banding artifacts, as illustrated in \cite{wang2016perceptual}.  Therefore, the advantage of  frequency-aware calibration in IQA metrics is not significant. To visually demonstrate the effectiveness of frequency-aware calibration, we present a comparison of the visual results from the model trained with original loss and frequency-aware calibration.

\begin{figure*}[t]
\scriptsize
\begin{center}
\begin{tabular}{cccccccc}
\hspace{-0.50cm}
\begin{adjustbox}{valign=t}
\begin{tabular}{cccccc}
\includegraphics[width=0.2447\textwidth]{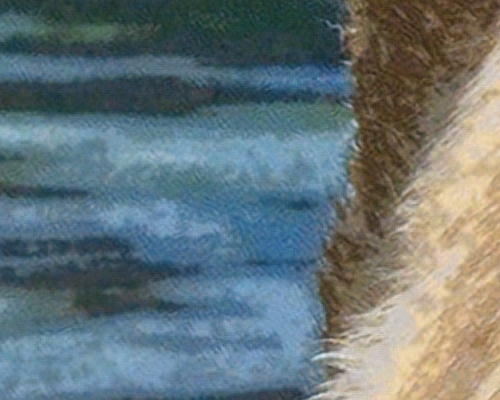} \hspace{-4.mm} &
\includegraphics[width=0.2447\textwidth]{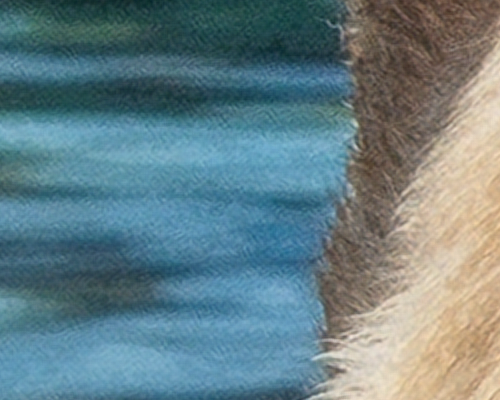} \hspace{-4.mm} &
\includegraphics[width=0.2447\textwidth]{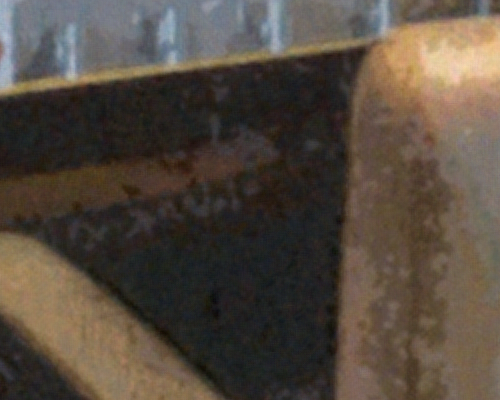} \hspace{-4.mm} &
\includegraphics[width=0.2447\textwidth]{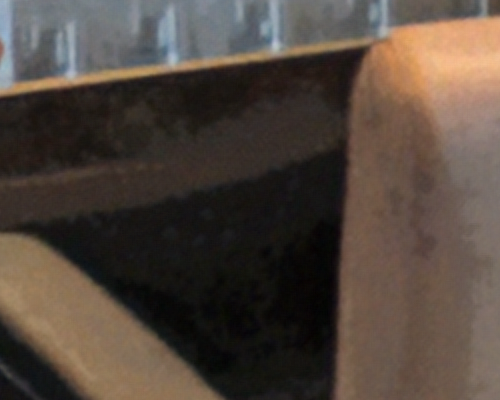} \hspace{-4.mm} &
\\ 
Orignal (UHDM 0047) \hspace{-4.mm} &
Frequency (UHDM 0047) \hspace{-4.mm} &
Orignal (UHDM 0071) \hspace{-4.mm} &
Frequency (UHDM 0071) \hspace{-4.mm} &
\\
\includegraphics[width=0.2447\textwidth]{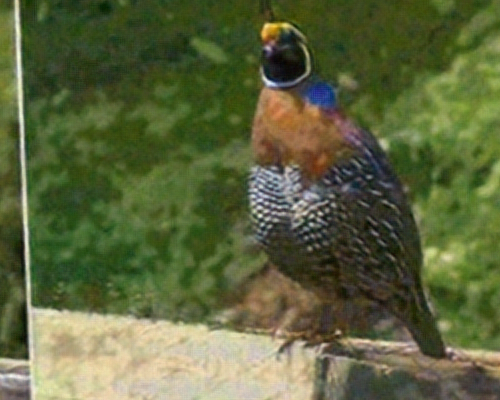} \hspace{-4.mm} &
\includegraphics[width=0.2447\textwidth]{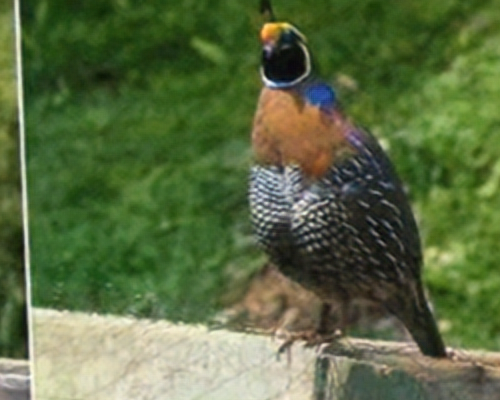} \hspace{-4.mm} &
\includegraphics[width=0.2447\textwidth]{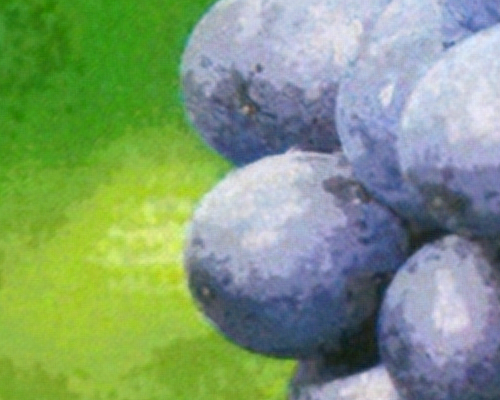} \hspace{-4.mm} &
\includegraphics[width=0.2447\textwidth]{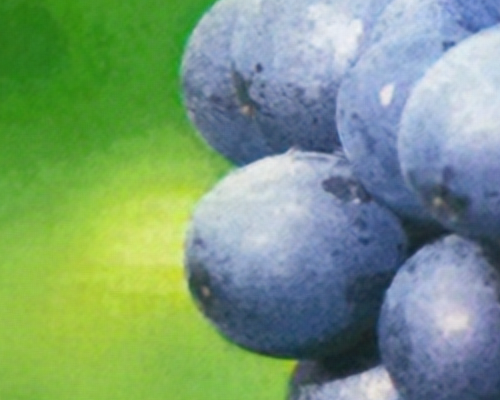} \hspace{-4.mm} &
\\ 
Orignal (UHDM 0047) \hspace{-4.mm} &
Frequency (UHDM 0047) \hspace{-4.mm} &
Orignal (UHDM 0071) \hspace{-4.mm} &
Frequency (UHDM 0071) \hspace{-4.mm} &
\\
\end{tabular}
\end{adjustbox}

\end{tabular}
\end{center}
\caption{Visualization of the effectiveness of frequency-aware calibration. "Orignal" are the results of model trained with original loss function, "Frequency" are the results of model trained with frequency-aware calibration, which significantly weakened banding artifacts.}
\label{fig:banding_more}
\end{figure*}

\begin{table*}[t]
\scriptsize
\begin{center}
\setlength{\tabcolsep}{4.7mm}
\begin{tabular}{l|c|c|c|c}
    \toprule
    \rowcolor{color3}  {Method} & {PSNR} & {SSIM} & {LPIPS} & Training epoch \\
    \midrule
    Ours (only learn the clipping bound)  & 21.0808 & 0.7626 & 0.3068 & 4\\
    {Omniquant (learn clipping bound and smoothing factor)}  & 21.0643 & 0.7613 & 0.3106 & 40\\
    \bottomrule
\end{tabular}
\end{center}
\caption{Comparison of our method vs. Omniquant. Experiments are conducted in 4bit, UHDM.}
\label{tab:omni}
\end{table*}

\begin{figure*}[t]
\scriptsize
\begin{center}
\begin{tabular}{cccccccc}
\hspace{-4.5mm}
\begin{adjustbox}{valign=t}
\begin{tabular}{cccccc}
\includegraphics[width=0.2447\textwidth]{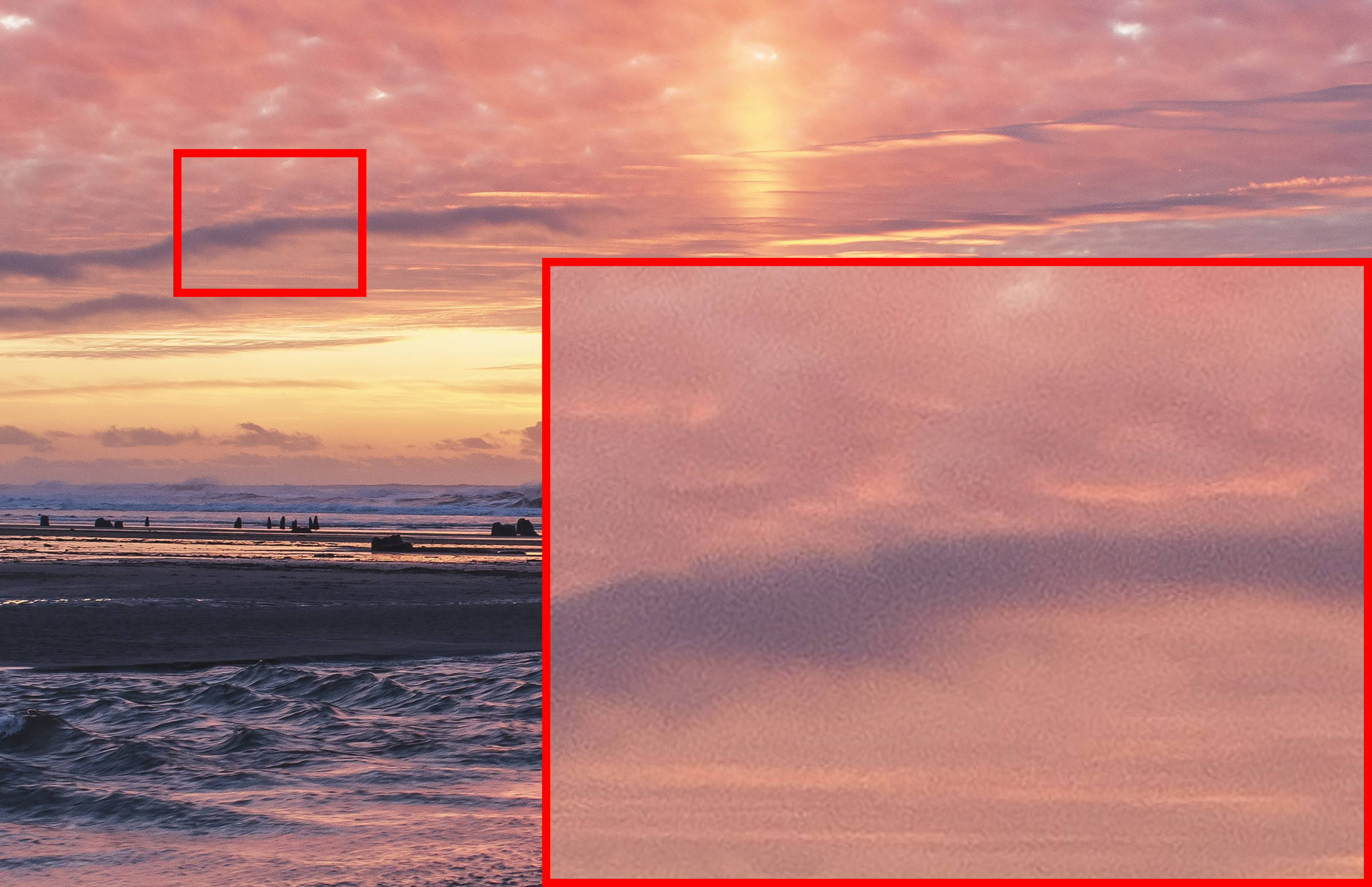} \hspace{-4.mm} &
\includegraphics[width=0.2447\textwidth]{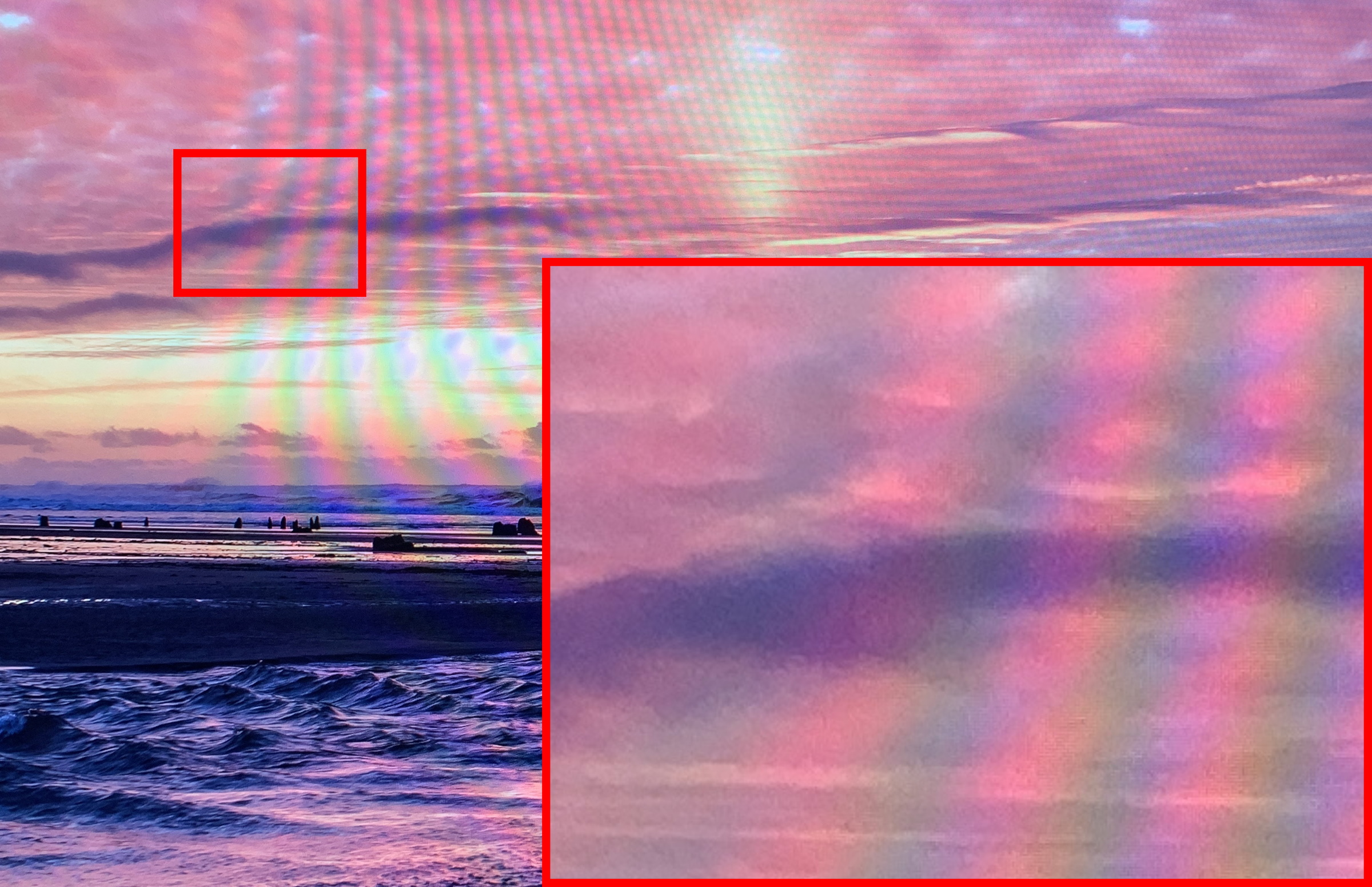} \hspace{-4.mm} &
\includegraphics[width=0.2447\textwidth]{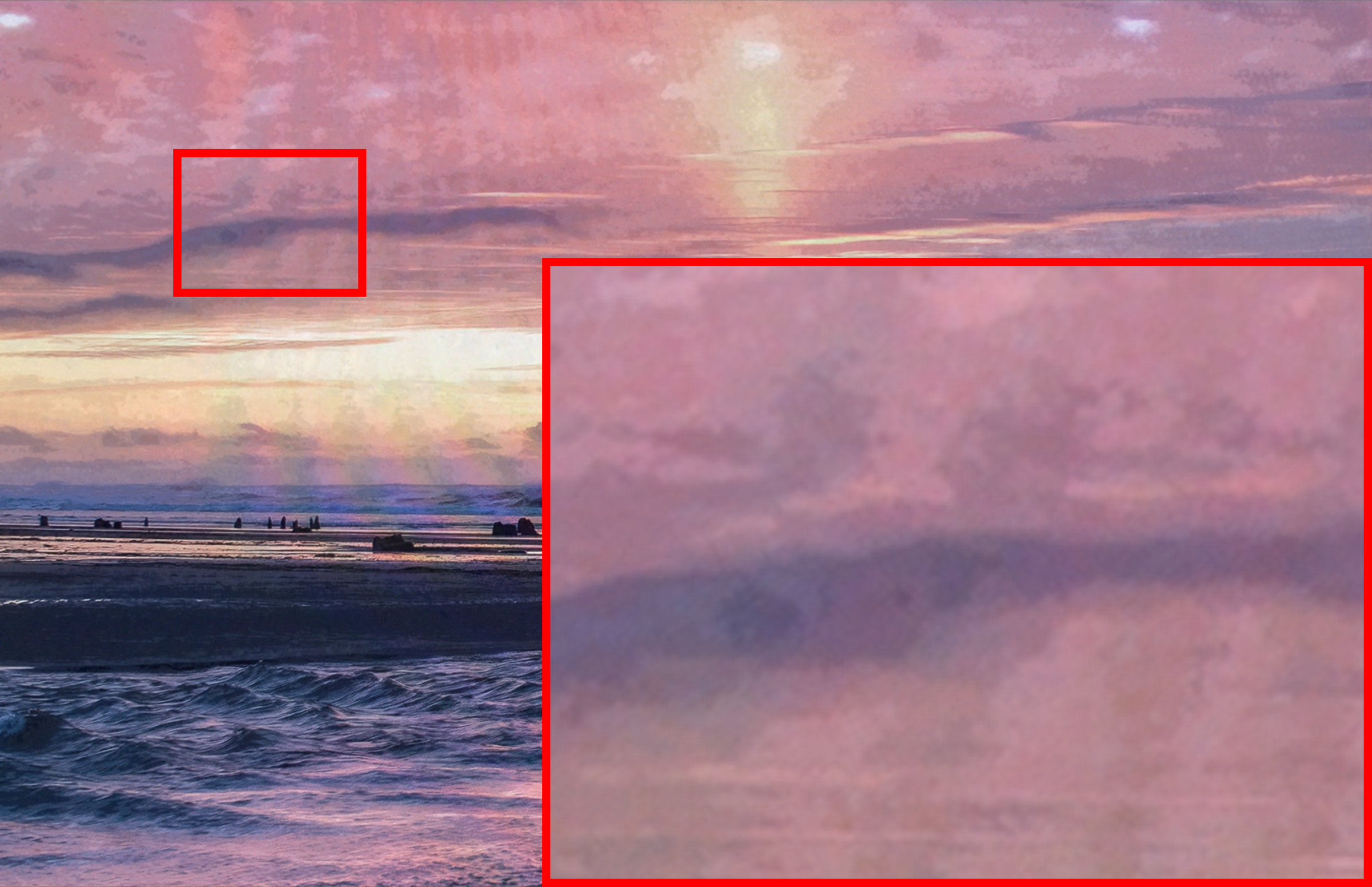} \hspace{-4.mm} &
\includegraphics[width=0.2447\textwidth]{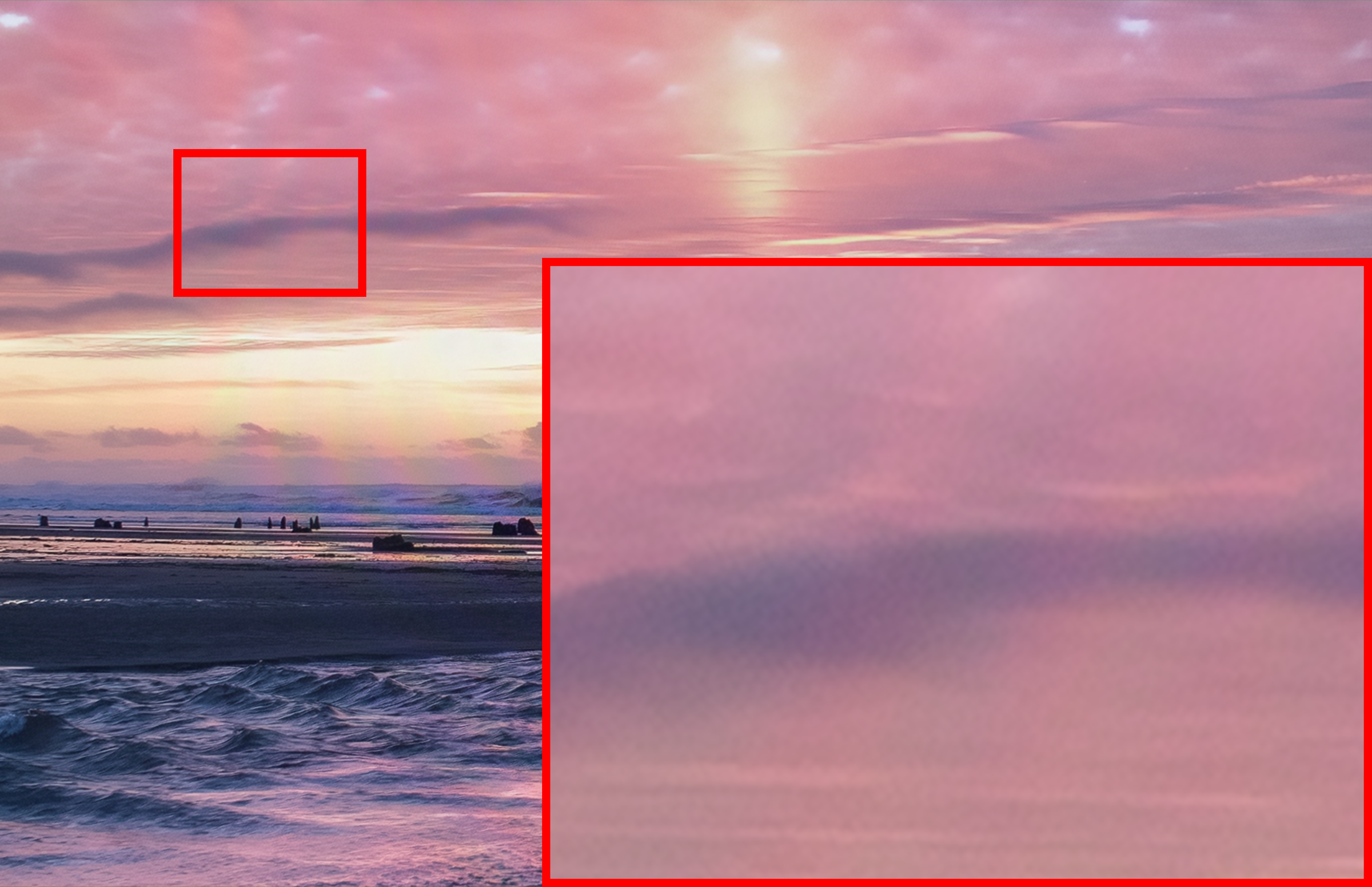} \hspace{-4.mm} &
\\ 
GT \hspace{-4.mm} &
Input \hspace{-4.mm} &
Original (banding) \hspace{-4.mm} &
Mid\&Low-Freq  \hspace{-4.mm} &
\\
\end{tabular}
\end{adjustbox}

\end{tabular}
\end{center}
\vspace{-4.mm}
\caption{Visualization of results under different calibration strategies. Original: training in the spatial domain suffers from banding artifacts under low-bit quantization. Mid\&Low-Freq: our frequency-aware calibration strategy leverages mid- and low-frequency.}
\vspace{-2.mm}
\label{fig:frequency}
\end{figure*}

It is evident that under the extremely low bit-width setting of 4-bit, the baseline model (using the original loss) still suffers from severe banding artifacts. In contrast, frequency-aware calibration significantly alleviates this issue.

\subsection{Implementation Details of the FP16 branch}
\label{sec:efficiency_analysis}
Our Mixed-Precision Weight Quantizer retains only 1\% of convolutional weights, creating extreme unstructured sparsity that is inherently unfriendly to parallel execution. We address this challenge with a specialized GPU kernel. We employ CSR-style per-channel indexing for sparse weight storage. The computation strategy tiles output rows across CUDA blocks and stripes threads along the column dimension, with each kernel serially traversing the sparse weights and accumulating contributions to the output tensor.

This design choice is justified by the extreme sparsity regime: for a typical 32×16×3×3 kernel, each output channel retains only 1-2 weight elements on average, so serial traversal is more efficient than parallel scheduling. This approach successfully exploits unstructured sparsity while incurring only 6\% GPU memory. On an RTX 4090 GPU, our fused CUDA implementation achieves a 2.3$\times$ end-to-end speedup over the FP16 model.

\subsection{Comprehensive Ablation Study Details}
\label{sec:app_ablation_details}

All ablation experiments use static quantization.

\textbf{Table 1 (a) Quantizer bound calculation:} All methods are evaluated without smoothing and without any training-based optimization. Weights are quantized using MinMax. For activation quantization: MinMax uses the absolute minimum and maximum; Percentile applies 99.9th-percentile clipping; Sample uses the sampling-based estimator proposed.  

\textbf{Table 1 (b) Smooth transformation:} In this setting, weights are quantized by MinMax, activations are quantized using the sampling-based estimator, and no training-based optimization is applied. Original applies no smoothing; Smooth (Raw) computes the smoothing factor from the absolute maximum activation; Smooth (Sample) computes the smoothing factor using the sampling-based estimator.

\textbf{Table 2 (a) Outlier handling method:} Baseline and Smooth (Sample) in Table 1 (b) share identical settings. Discard Outliers truncates weight outliers; Store Random randomly stores 1\% of weights in FP16; Store Outlier stores the largest 0.5\% and smallest 0.5\% of weights.

\textbf{Table 2 (b) Outliers Proportion:} 0.50\% is identical to Store Outlier in Table 2 (a); 0.25\% stores the largest 0.25\% and smallest 0.25\% of weights (all other settings identical to the 0.50\% case); 1\% stores the largest 1\% and smallest 1\% of weight elements (others identical to the 0.50\%).

\textbf{Table 3 (a) Calibration strategy:}  No Calibration uses the same settings as Store Outlier in Table 2 (a). Original Image applies $L_1$ \& $L_p$ calibration on the original images (on top of No Calibration). High-Freq performs calibration using high-frequency  image components. Mid\&Low-Freq performs calibration using mid- and low-frequency components.

\textbf{Table 3 (b) Frequency extraction level:}  For level = 3, the settings are identical to Mid\&Low-Freq in Table 3 (a). For level = 1 and level = 5, we use 1 and 5 convolution kernels, respectively, to extract frequency components; all other settings are the same as for level = 3.

\subsection{Comparison with Related Methods}
\label{sec:app_related_methods}

\subsubsection{Comparison with OmniQuant}

Both the outlier-aware quantizer and OmniQuant~\cite{OmniQuant} follow the same high-level smoothing-and-clipping scheme. However, OmniQuant~\cite{OmniQuant} is designed to quantize large language models (LLMs), whereas the outlier-aware quantizer targets an image demoir\'eing model composed mainly of convolutional modules. Consequently, the two approaches differ substantially in their specific design choices.

\textbf{(1) Different clipping targets.} OmniQuant~\cite{OmniQuant} employs learnable weight clipping in LWC. By contrast, in convolutional modules the weights are confined to small kernels and are relatively few per layer compared with the volume of activations. Truncating weight outliers can therefore cause disproportionate accuracy loss. Therefore, in the outlier-aware quantizer we clip only the activation quantization bounds rather than clipping the weights.

\textbf{(2) Choice of learnable parameters.} OmniQuant~\cite{OmniQuant} treats both the smoothing factor and the weight clipping bounds as learnable. In our outlier-aware quantization, only the activation clipping bound is learnable; the smoothing factor is initialized using a sampling-based estimator and then kept fixed in the training process. Because the rounding operation in quantization can destabilize training, we observe that our design outperforms the alternative that makes both the smoothing factor and the weight clipping bounds learnable, as the results shown in Tab.~\ref{tab:omni}.

Furthermore, by initializing the smoothing factor to a good value via sampling and reducing the number of learnable parameters, our method requires only 4 training epochs, whereas the configuration with learnable smoothing factor and clipping bounds typically requires 40 epochs. Our approach substantially reduces the training time.

\subsubsection{Comparison with Mixed-Precision Methods}

Many prior methods adopt mixed-precision quantization. For example, LLM.int8~\cite{dettmers2022llmint88bitmatrixmultiplication} processes a very small number of outlier feature dimensions in FP16, and SqueezeLLM~\cite{kim2024squeezellmdenseandsparsequantization} stores a small subset of weights in a sparse format. However, few approaches are designed for convolutions. Because outlier weights in convolution kernels occur sporadically, existing coarse-grained mixed-precision schemes are not well suited, and directly converting convolutions into sparse matrix multiplications introduces additional overhead. To address this, we implement a dedicated kernel for the FP16 branch that parallelizes processing of FP16 weights at the element level.

\subsection{Visualization of Different Calibration Strategies}

Results under different calibration strategies are shown in Fig.~\ref{fig:frequency}. Training directly in the spatial domain (Original) causes the quantized demoir\'eing model to exhibit severe banding artifacts, which degrades output quality. Our Frequency-Aware Calibration Strategy successfully mitigates banding by optimizing specific frequency components, achieving results close to the ground truth.

\section{Explanations for Checklist}
\label{sec:app_checklist}
\subsection{Limitations}
Despite achieving strong results, QuantDemoire still has several limitations. First, our method requires a representative calibration set; without calibration data, the sampling-based range estimation and frequency-aware calibration cannot be reliably performed. Second, the sampling strategy is most effective when activations and weights contain sparse extreme outliers around a dominant distribution. For layers with highly irregular or strongly multi-modal distributions, the sampling rates may require additional tuning. Third, the current mixed-precision branch relies on a custom CUDA kernel evaluated on an RTX 4090 GPU, and deployment on mobile NPUs or other accelerators may require hardware-specific engineering.

\subsection{Broader Impacts}
QuantDemoire reduces the memory, computation, and energy cost of image demoir\'eing models, facilitating their deployment on edge devices such as cameras, drones, and portable imaging systems. This can improve the accessibility of high-quality image restoration under limited computational resources. At the same time, more efficient visual restoration may also be used in privacy-sensitive applications such as surveillance or large-scale data collection, and quantization artifacts under distribution shifts may affect downstream visual analysis. We therefore encourage careful validation on target devices and datasets before deployment.


\begin{figure*}[!t]
\vspace{20.mm}
\scriptsize
\centering

\caption{Visual comparison on the UHDM~\cite{yu2022towards} dataset.}
\label{fig:more-visual-1}
\vspace{20.mm}
\end{figure*}

\begin{figure*}[!t]
\vspace{20.mm}
\scriptsize
\centering
%
\caption{Visual comparison on the LCDMoir\'e~\cite{yuan2019aim} dataset.}
\label{fig:more-visual-2}
\vspace{20.mm}
\end{figure*}

\begin{figure*}[!t]
\vspace{30.mm}
\scriptsize
\centering
%
\caption{More distribution of activation (Part 1).}
\label{fig:act_distribution1}
\vspace{30.mm}
\end{figure*}

\begin{figure*}[!t]
\vspace{17.mm}
\scriptsize
\centering
%
\caption{More distribution of activation (Part 2).}
\label{fig:act_distribution2}
\end{figure*}

\begin{figure*}[!t]
\scriptsize
\centering
%
\caption{More distribution of weight (Part 1).}
\label{fig:weight_distribution1}
\vspace{17.mm}
\end{figure*}

\begin{figure*}[!t]
\vspace{26.mm}
\scriptsize
\centering
%
\end{adjustbox}
\\

\end{tabular}
\caption{More distribution of weight (Part 2).}
\label{fig:weight_distribution2}
\vspace{26.mm}
\end{figure*}

\end{document}